\documentclass[10pt,twocolumn,letterpaper]{article}

\usepackage[pagenumbers]{cvpr} 

\usepackage[dvipsnames]{xcolor}

\usepackage{graphicx}
\usepackage{amsmath}
\usepackage{amssymb}
\usepackage{booktabs}

\usepackage{mathtools}
\usepackage[nolist]{acronym} 

\usepackage{bm}
\usepackage{tablefootnote}
\usepackage{soul}

\begin{acronym}[PHOENIX14\textbf{T} ]

\acrodefplural{rnn}[RNNs]{Recurrent Neural Networks}
\acrodefplural{cnn}[CNNs]{Convolutional Neural Networks}
\acrodefplural{hmm}[HMMs]{Hidden Markov Models}
\acrodefplural{gru}[GRUs]{Gated Recurrent Units}
\acrodefplural{crf}[CRFs]{Conditional Random Fields}
\acrodefplural{gan}[GANs]{Generative Adversarial Networks}
\acrodefplural{gpu}[GPUs]{Graphic Processing Units}

\acrodefplural{mdn}[MDNs]{Mixture Density Networks}

\acro{aslt}[ASLT]{American Sign Language Translation}
\acro{aslp}[ASLP]{American Sign Language Production}
\acro{bsl}[BSL]{British Sign Language}
\acro{bleu}[BLEU]{Bilingual Evaluation Understudy}
\acro{blstm}[BLSTM]{Bidirectional Long Short-Term Memory}
\acro{bslcpt}[BSLCP\textbf{T}]{BSL Corpus \textbf{T}}
\acro{bslp}[BSLP]{Bilingual Sign Language Production}
\acro{cnn}[CNN]{Convolutional Neural Network}
\acro{crf}[CRF]{Conditional Random Field}
\acro{cslr}[CSLR]{Continuous Sign Language Recognition}
\acro{ctc}[CTC]{Connectionist Temporal Classification}
\acro{c4a}[C4A]{Content4All}
\acro{dl}[DL]{Deep Learning}
\acro{dgs}[DGS]{German Sign Language - Deutsche Gebärdensprache}
\acro{dsgs}[DSGS]{Swiss German Sign Language - Deutschschweizer Geb\"ardensprache}
\acro{dtw}[DTW]{Dynamic Time Warping}
\acro{fc}[FC]{Fully Connected}
\acro{ff}[FF]{Feed Forward}
\acro{gan}[GAN]{Generative Adversarial Network}
\acro{gpu}[GPU]{Graphics Processing Unit}
\acro{gru}[GRU]{Gated Recurrent Unit}
\acro{gslp}[GSLP]{German Sign Language Production}
\acro{gtpt}[G2PT]{Gloss to Pose Transformer}
\acro{hmm}[HMM]{Hidden Markov Model}
\acro{hpe}[HPE]{Hand Pose Enhancer}
\acro{isl}[ISL]{Irish Sign Language}
\acro{lstm}[LSTM]{Long Short-Term Memory}
\acro{mha}[MHA]{Multi-Headed Attention}
\acro{mtc}[MTC]{Monocular Total Capture}
\acro{mse}[MSE]{Mean Squared Error}
\acro{mslp}[MSLP]{Multilingual Sign Language Production}
\acro{mdn}[MDN]{Mixture Density Network}
\acro{mdgs}[mDGS]{meineDGS}
\acro{mdgsv}[mDGS-V]{meineDGS-Variants}
\acro{mdgst}[mDGS\textbf{T}]{meineDGS\textbf{T}}
\acro{mdgsth}[mDGS\textbf{T}-\textbf{H}]{meineDGS\textbf{T}-\textbf{HARD}}
\acro{mdgste}[mDGS\textbf{T}-\textbf{E}]{meineDGS\textbf{T}-\textbf{EASY}}

\acro{nmt}[NMT]{Neural Machine Translation}
\acro{nlp}[NLP]{Natural Language Processing}
\acro{ph12}[PHOENIX12]{RWTH-PHOENIX-Weather-2012}
\acro{ph14}[PHOENIX14]{RWTH-PHOENIX-Weather-2014}
\acro{ph14t}[PHOENIX14\textbf{T}]{RWTH-PHOENIX-Weather-2014\textbf{T}}
\acro{pof}[POF]{Part Orientation Field}
\acro{paf}[PAF]{Part Affinity Field}
\acro{pttt}[P2TT]{Pose to Text Transformer}
\acro{relu}[RELU]{Rectified Linear Units}
\acro{rnn}[RNN]{Recurrent Neural Network}
\acro{rouge}[ROUGE]{Recall-Oriented Understudy for Gisting Evaluation}
\acro{sgd}[SGD]{Stochastic Gradient Descent}
\acro{sla}[SLA]{Sign Language Assessment}
\acro{slr}[SLR]{Sign Language Recognition}
\acro{slt}[SLT]{Sign Language Translation}
\acro{slp}[SLP]{Sign Language Production}
\acro{smt}[SMT]{Statistical Machine Translation}

\acro{ttgt}[T2GT]{Text to Gloss Transformer}
\acro{ttpt}[T2PT]{Text to Pose Transformer}
\acro{ttp}[T2P]{Text to Pose}
\acro{ttgtp}[T2G2P]{Text to Gloss to Pose}
\acro{tts}[TTS]{Text to Speech}
\acro{wer}[WER]{Word Error Rate}

\end{acronym} 
\usepackage[linesnumbered,lined,boxed,commentsnumbered]{algorithm2e}
\newcommand{\ourLLMmethodName}{\textsc{SignLLM}}
\newcommand{\ourDataName}{\textsc{Prompt2Sign}}

\newcommand{\greencheck}{{\color{lightGreen}\checkmark}}
\usepackage{pifont}
\newcommand{\xmark}{\color{red}\ding{55}}%

\usepackage{caption}
\usepackage[nolist]{acronym}

\usepackage{dblfloatfix}
\usepackage[dvipsnames]{xcolor}

\usepackage{multirow}

\usepackage{enumitem,xcolor}
\newlist{coloritemize}{itemize}{1}
\setlist[coloritemize]{label=\textcolor{itemizecolor}{\textbullet},font=\bfseries\color{itemizecolor}}
\colorlet{itemizecolor}{red}
\usepackage[accsupp]{axessibility}  
\usepackage{titletoc} 
\usepackage{minitoc}

\usepackage{tabularx} 
\usepackage{colortbl}
\usepackage{wrapfig}
\usepackage[title]{appendix}
\usepackage{inconsolata}

\definecolor{cvprblue}{rgb}{0.21,0.49,0.74}

\definecolor{backgroundblue}{RGB}{255,255,255} 
\definecolor{backgroundgray}{RGB}{255,255,255} 
\definecolor{cancelgray}{RGB}{0,0,0} 

\definecolor{backgroundgreen}{RGB}{213,232,212} 
\definecolor{redongray}{RGB}{192,0,0} 
\definecolor{blueongray}{RGB}{0,176,240} 
\definecolor{lightGreen}{RGB}{0,210,0} 
\definecolor{sgreen}{RGB}{30, 150, 30} 
\definecolor{arXivGreen}{RGB}{34,139,34} 
\usepackage[pagebackref,breaklinks,colorlinks,allcolors=cvprblue]{hyperref}
\hypersetup{urlcolor=magenta}

\usepackage{orcidlink}

\def\paperID{12664} 
\def\confName{ICCV}
\def\confYear{2025}

\begin{document}

\title{SignLLM: Sign Language Production Large Language Models}


\author{
Sen Fang\textsuperscript{\rm 1}, Chen Chen\textsuperscript{\rm 2}, Lei Wang\textsuperscript{\rm 3,4}, Ce Zheng\textsuperscript{\rm 5}, Chunyu Sui\textsuperscript{\rm 6}, Yapeng Tian\textsuperscript{\rm 7}
\\
{\small \textsuperscript{\rm 1} Rutgers University, \textsuperscript{\rm 2} University of Central Florida, \textsuperscript{\rm 3} Australian National University, \textsuperscript{\rm 4} Data61/CSIRO} 
\\
{\small \textsuperscript{\rm 5} Carnegie Mellon University, \textsuperscript{\rm 6} Columbia University, \textsuperscript{\rm 7} The University of Texas at Dallas}
\\
{\normalsize\url{https://signllm.github.io}}
\vspace{-16pt}
}

\twocolumn[{%
\renewcommand\twocolumn[1][]{#1}%
\maketitle

\DeclareRobustCommand*{\RaiseBoxByDepth}{%
    \raisebox{-0.2\height}%
}

\begin{center}
    \centering
    \vspace{-6pt}
    \includegraphics[width=.99\textwidth]{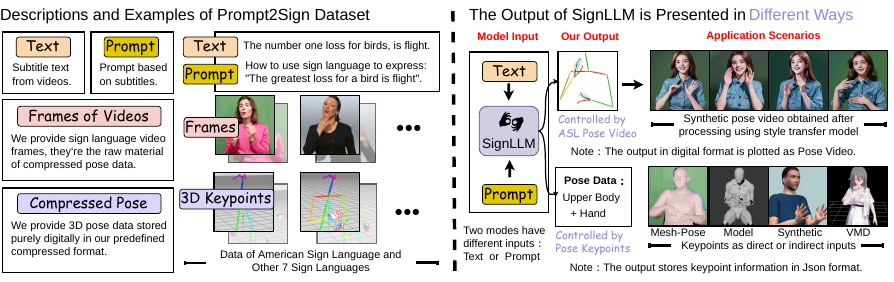}
    \vspace{-8pt}
    \captionof{figure}{\textbf{Overview:} (Left) Major components (\eg, 
    \href{none}{\RaiseBoxByDepth{\includegraphics[height=8pt]{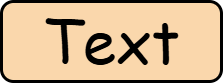}}}, 
    \href{none}{\RaiseBoxByDepth{\includegraphics[height=8pt]{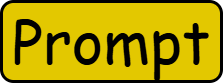}}}, 
    \href{none}{\RaiseBoxByDepth{\includegraphics[height=8pt]{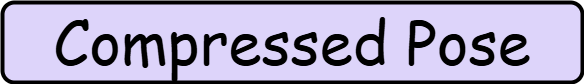}}}, \etc) of our \ourDataName{} dataset. Compressed Pose is reprocessed pose data that is suitable for training, we use public sign language videos to produce compressed pose data in our predefined format;
    (Right) Our proposed \ourLLMmethodName{} aims to generate sign language poses for digital human or avatar generation \cite{chen2023executing,cai2023smplerx,zwitserlood2004synthetic,zhang2023adding}. 
    }
    \label{fig:cover}
\end{center}%
}]
\begin{abstract}
In this paper, we propose \textbf{\ourLLMmethodName{}}, a multilingual \ac{slp} large language model, which includes two novel multilingual SLP modes MLSF and Prompt2LangGloss that allow sign language gestures generation from query texts input and question-style prompts input respectively. Both modes can use a new RL loss based on reinforcement learning and a new RL module named Priority Learning Channel. These RL components can accelerate the training by enhancing the model's capability to sample high-quality data.
To train \ourLLMmethodName{}, we introduce \textbf{\ourDataName{}}, a comprehensive multilingual sign language dataset, which builds from public data, including American Sign Language (ASL) and seven others. This dataset standardizes information by extracting pose information from sign language videos into a unified compressed format. We extensively evaluate \ourLLMmethodName{}, demonstrating that our model achieves state-of-the-art performance on SLP tasks across eight sign languages.

\end{abstract}    

\section{Introduction}
\label{sec:intro}

\acf{slp} aims to synthesize human-like sign avatars from text inputs.
Deep learning-based SLP approaches \cite{saunders2020progressive,saunders2021continuous,huang2021towards,saunders2021mixed,saunders2021skeletal} typically involve sequential steps from text to gloss (\ie, a type of textual vocabulary representing gestures or postures), gloss to pose \cite{carreira-2017-i3d,MediaPipe},
and finally rendering pose videos into more engaging human-like avatar videos. These processes are complex and challenging to simplify, making sign language data acquisition and processing difficult. This challenge has significantly dampened researchers' enthusiasm and progress over a considerable period, with the majority of studies in the past decade relying on a German sign language (GSL) dataset named PHOENIX14T \cite{forster2012rwth, koller15:cslr} for \textbf{S}ign \textbf{L}anguage \textbf{P}roduction, \textbf{R}ecognition, and \textbf{T}ranslation tasks (SLP, SLR and SLT). 
Recent work \cite{slt-how2sign-wicv2023,fang2024signdiffdiffusionmodelsamerican,Bohacek_2022_WACV} based on the American sign language (ASL) \cite{duarte2021how2sign,shi-etal-2022-openASL} and other lesser-known languages \cite{Duarte_2021_how2sign,RWTH-PHOENIX-Weather-2014,muller-etal-2023-findings,Ronchetti2016,Ham21,9210578} are relatively rare.

The existing mainstream datasets \cite{SLTranslation,duarte2021how2sign} have significantly advanced the field. 
However, as time progresses, their limitations become increasingly apparent:
(1) These existing datasets consist of different format files, including images, Glosses, subtitles, \etc. These images are not easy to be directly trained. Due to redundant information in images that makes it difficult for models to learn essential pose information, training video-level SLP becomes particularly challenging.
A way to reduce redundant information is to distill the gesture/posture information into text/npy/json for training. But different datasets pose extraction methods \cite{7298594,guler2018densepose,smpl,qiu2021dense,carreira-2017-i3d,MediaPipe} are different. This limitation hampers a specific format model's ability to use data from other sign languages. (2) Manual annotation for gloss is labor-intensive and time-consuming. (3) Because some videos are obtained from professionals and reprocessed into different formats, scaling up the dataset becomes exceedingly challenging.
These limitations collectively impede the development and training of more advanced models. 

To solve these issues, we introduce \textbf{\ourDataName{}}, a new multilingual dataset focusing on upper body movements in large-scale signers. The dataset overview is shown in Fig.~\ref{fig:cover} (Left), showcasing prompts, video subtitles, and files containing digital keypoints information. To create this dataset, we first process the videos using OpenPose \cite{cao2018openpose} to standardize pose information in each frame. Storing keypoints information in our predefined compressed format (\href{none}{\raisebox{-0.2\height}{\includegraphics[height=8pt]{fig/icon/pose.drawio.png}}}, as shown in Fig. \ref{fig:cover}) can reduce redundancy and facilitate training with seq2seq and text2text models. Subsequently, we reduce reliance on manual annotations by auto-creating prompt words to improve cost-effectiveness. Finally, we improve the processing level of automation for the tools, making the tools highly efficient and lightweight, requiring no additional model loading to process data (\ie, solving the difficulty in manual preprocessing and data collection above).
Our new \ourDataName{} dataset is sourced from publicly available sign language datasets and videos on the Internet, 
covering eight sign languages,
making it a comprehensive multilingual sign language dataset. More information is in Table \ref{tab:dataset_statistics} and supplementary materials.

Meanwhile, we recognize that existing models \cite{wang2018video,wang2018high,chan2019everybody,saunders2020progressive,stoll2020text2sign,saunders2021mixed} need improvement because training models with our new dataset brings new challenges:
(1) Different sign language data cannot usually be trained simultaneously due to text-posture correspondence differences in different sign languages. (2) Handling more languages and a larger dataset results in slow and challenging training processes, with downloading, storing, and data loading difficulties. It is necessary to explore high-speed training methods.
(3) The existing model structure cannot grasp more languages and understand more complex natural human conversational inputs.  
So, we need to explore the aspects overlooked by previous studies, such as multilingual SLP, efficient training, and the ability to understand prompts.

To overcome these challenges, we introduce \textbf{\ourLLMmethodName{}}, a large multilingual \acf{slp} model developed based on our \ourDataName{} dataset. It can produce the sign language representation of eight languages from texts or prompts. Our \ourLLMmethodName{} has two distinct modes: \href{none}{(i)} Multi-Language Switching Framework (MLSF), which allows multiple sign languages production in parallel by dynamically adding encoder-decoder groups. \href{none}{(ii)} Prompt2LangGloss, allowing \ourLLMmethodName{} to support static single-set encoder-decoder generation. Fig.~\ref{fig:cover} (Right) shows our model inputs and outputs, \textit{``thank you''} is the input of mode (i), and \textit{``how to sign `thank you' in ASL?''} is the input of mode (ii).
Two multilingual SLP modes deal with different use cases: The Multi-Language Switching Framework (MLSF) is an efficient mode without semantic confusion, like a dictionary/drawer; The Prompt2LangGloss is a user-friendly mode, like a LLM, it aims to understand complex natural language input. 
To address the problem of extended training time caused by more languages and a larger dataset, we utilize the Reinforcement Learning (RL) Loss concepts to quantify the quality of each training batch and prioritize valuable batches through the Priority Learning Channel.


We conduct extensive experiments and detailed ablation studies. The results validate the superior performance of our \textbf{\ourLLMmethodName{}} over baseline approaches \cite{wang2018video,wang2018high,chan2019everybody,saunders2020progressive,stoll2020text2sign,saunders2021mixed,ebling2016automatic,8778347} on the subsets in eight sign languages. 
The contributions of this paper can be summarized as follows.
\begin{itemize}
    \item A comprehensive multilingual sign language dataset, named \ourDataName{}, featuring an expanded vocabulary and covering eight languages, is introduced. It is designed for broader seq2seq compatibility.
    \item A large multilingual \acf{slp} model with two distinct modes—MLSF for handling text query inputs and Prompt2LangGloss for processing question-style prompts—is proposed. Our method, \ourLLMmethodName{}, achieves state-of-the-art performance across eight sign languages in \ac{slp} tasks. 
    \item We present a novel reinforcement learning-based loss function, along with a functional module named Priority Learning Channel (PLC), as a training strategy for sign language models, designed to reduce training time and computational costs.  
\end{itemize}



\section{Related Work} \label{sec:related_work}
%
\textbf{Sign Language Production.}
In recent years, the field of sign language research has primarily focused on \acf{slr} \cite{cui2017recurrent,grobel1997isolated,koller2020quantitative,kadir2004minimal,cooper2007large,koller2015continuous} and \acf{slt} \cite{camgoz2018neural,camgoz2020sign,ko2019neural,Bohacek_2022_WACV} based on deep learning. 
Due to a lack of suitable, high-quality datasets for ASL, deep learning researchers conducted their research \cite{saunders2020progressive,saunders2021continuous,huang2021towards,saunders2021mixed,saunders2021skeletal} based on a GSL weather theme dataset, released in 2012 \cite{forster2012rwth, koller15:cslr}.
As previously mentioned, the data processing involved in sign language research is highly complex. Even with the release of a large-scale ASL dataset
in 2021 \cite{duarte2021how2sign}, work focused on ASL-related themes based on it has not emerged quickly, as existing work is not easily transferable. The situation is worse for minority languages.



\noindent\textbf{Large Language Models.}
LLMs refer to giant transformer models trained on extensive textual data, exhibit capabilities in understanding natural language and addressing complex tasks \cite{Shanahan-arxiv-2022-Talking,Brown-NeurIPS-2020-Language,Chowdhery-arxiv-2022-PaLM,Taylor-arxiv-2022-Galactica,Touvron-arxiv-2023-LLaMA}. Sign language is a visual language, theoretically different from language models. However, most current work uses text2text and seq2seq models \cite{raffel2020exploring,xue2020mt5,JoeyNMT,paszke2017automatic}, converting key points/dense maps/grid poses into sequences for training, as opposed to directly training images. Hence, viewing the core process of \ac{slp}, text2pose, as a language model is justifiable.
Extensive research indicates that an increase in parameters or data volume \cite{Hoffmann-arxiv-2022-Training, Kaplan-arxiv-2020-Scaling} significantly enhances the abilities of LLMs \cite{radford-blog-2019-language,Brown-NeurIPS-2020-Language,Chowdhery-arxiv-2022-PaLM}. There are more than a hundred sign languages in the world, most of which have datasets in video form. Therefore, conducting advanced research to address the anticipated surge in data volume in the future is of paramount importance. 
In this work, we aim to enable the model to generate sign languages across diverse linguistic backgrounds, ensuring its adaptability to our new dataset.


\section{Our Benchmark: Prompt2Sign} \label{sec:prompt2sign}


\noindent\textbf{Existing Datasets Weaknesses.} The main shortcoming of previous work is the lack of unified data storage formats, 
when there is a mismatch between these models in \acf{slp} \cite{wang2018video,wang2018high,chan2019everybody,saunders2020progressive,stoll2020text2sign,saunders2021mixed} and \acf{slt} \cite{camgoz2018neural,camgoz2020sign,ko2019neural,Bohacek_2022_WACV}, it can lead to complex challenges: (1) The results of the SLT model are difficult to use as training data for the SLP model directly due to format incompatibility (\eg, \cite{slt-how2sign-wicv2023,cai2023smplerx} \& \cite{saunders2020adversarial,Saunders_2022_CVPR})
(2) The results of the SLP model are difficult to use as input for the SLT model (evaluation experiment needs, \eg, \cite{cai2023smplerx} \& \cite{Bohacek_2022_WACV}). (3) The output of the SLP model is not suitable as input for most style transfer models (\eg, \cite{chan2019everybody,wei2020gac,zhou2019dance}, the researchers have to train a pose2video model themselves). Therefore, we create a standardized dataset to address data collection, utilization, and storage challenges.

\noindent\textbf{Data Collection.}\label{paragraph:prompt2sign_collection}
Our data collection process, in contrast to previous methods, includes the following steps: (1) downloading sign language videos in specific languages from the Internet and public datasets \cite{Duarte_2021_how2sign, RWTH-PHOENIX-Weather-2014, muller-etal-2023-findings, Ronchetti2016, Ham21, 9210578}; (2) editing and aligning these videos; (3) extracting 2D keypoints from each video frame using OpenPose \cite{cao2018openpose} and saving them as JSON files; (4) calculating the 3D poses, which are stored in a predefined compressed data format. See further details below and in the \textit{supplementary materials}.

\noindent\textbf{Dataset Range.} 
We choose How2Sign \cite{Duarte_2021_how2sign}, PHOENIX-14T \cite{RWTH-PHOENIX-Weather-2014}, KSL-guide \cite{Ham21}, Signsuisse \cite{muller-etal-2023-findings} (contains 3 languages), LSA64 \cite{Ronchetti2016}, AuTSL \cite{9210578},
and some of the more popular works were not considered due to their limited accessibility and potential usage restrictions. Additionally, some available multilingual datasets \cite{gueuwou2023jwsign,yin2022mlslt,matthes2012dicta,hilzensauer2015multilingual} may not possess the same level of comprehensiveness as ours. For example, \cite{gueuwou2023jwsign} and \cite{yin2022mlslt} translate two types of sign language videos into spoken language (SLT), while our work is from spoken language to videos (SLP). We aim to establish a robust multilingual SLP method with a dataset that supports a broader range of application scenarios. For example, \cite{gueuwou2023jwsign} is limited to Bible translation, and \cite{yin2022mlslt} focuses solely on cross-lingual SLT. In contrast, our work covers a broader range of scenarios and comprehensively addresses the SLP task. Additionally, some datasets, such as \cite{matthes2012dicta} and \cite{hilzensauer2015multilingual}, are restricted to multilingual dictionary forms. 

\begin{table}[t]
\centering
\vspace{6pt}
\resizebox{0.48\textwidth}{!}{
\begin{tabular}{lcccccccc}
\toprule
\midrule
\textbf{Subset} & \textbf{ASL} & \textbf{GSL} & \textbf{DSGS} & \textbf{LSF-CH} & \textbf{LIS-CH} & \textbf{LSA} & \textbf{KSL} & \textbf{TSL} \\ \hline
Train & 31,047 & 7,096 & 8,043 & 5,672 & 2,254 & 2,400 & 700 & 28,142 \\ \hline
Dev & 1,739 & 519 & 500 & 500 & 250 & 400 & 300 & 4,418 \\ \hline
Test & 2,343 & 642 & 500 & 250 & 250 & 400 & 200 & 3,742 \\ 
\midrule
\bottomrule
\end{tabular}
}
\vspace{-6pt}
\caption{\textbf{Dataset Statistics} \cite{Duarte_2021_how2sign,RWTH-PHOENIX-Weather-2014,muller-etal-2023-findings,Ronchetti2016,Ham21,9210578, moryossef2021datasets, yang2019korean} \textbf{:} The number of video clips in their train, dev, test set, respectively. Languages included: American (ASL), German (GSL, Alias DGS), Swiss German (DSGS), French Sign Language of Switzerland (LSF-CH), Italian Sign Language of Switzerland (LIS-CH), Argentine (Lengua de Señas Argentina, LSA), Korean (KSL), and Turkish (TSL). ``Sign Language'' is omitted in some full names.
}
\label{tab:dataset_statistics}
\vspace{-17pt}
\end{table}




\begin{figure*}[t!]
    \centering
    \includegraphics[width=1\linewidth]{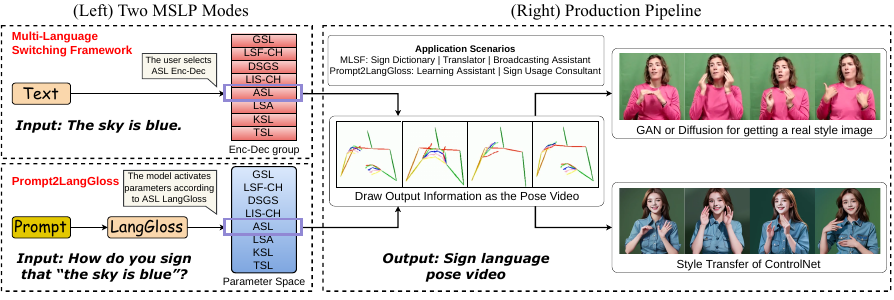}
    \vspace{-19pt}
    \caption{(Left) MLSF contains parallel Enc-Dec groups (\ie, \textit{Text2Pose $\times$ number of languages}), the Prompt2LangGloss adds a language attribute marker at the gloss channel (\ie, \textit{Text2Gloss2Pose $\rightarrow$ Prompt2LangGloss2Pose}). (Right) The output of \ourLLMmethodName{} can be converted into a skeletal pose video, which can then be rendered into a realistic human appearance by vid2vid models \cite{NEURIPS2022_ec795aea,chan2019everybody,wei2020gac,zhou2019dance,zhang2023adding}.} 
    \label{fig:Model_Overview}
    \vspace{-12pt}
\end{figure*}%

\noindent\textbf{Unified Compression of Pose Data.}
\label{paragraph:prompt2sign_format}
Building on previous work \cite{zelinka2020neural,saunders2020progressive,Saunders_2022_CVPR}, we develop a three-step tool for standardizing data processing. 
The tool is highly efficient and lightweight, requiring no additional model loading and supporting large-scale data processing. Furthermore, we optimize it specifically for sign language data processing (e.g., removing unnecessary leg movement computations and integrating these optimizations into \ourDataName{}'s pipeline).
The main steps of Data Collection can be visualized as: \href{none}{\raisebox{-0.2\height}{\includegraphics[height=8pt]{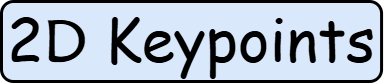}}} to \href{none}{\raisebox{-0.2\height}{\includegraphics[height=8pt]{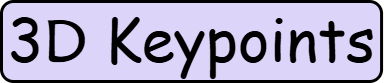}}} to \href{none}{\raisebox{-0.2\height}{\includegraphics[height=8pt]{fig/icon/pose.drawio.png}}}.
Among all the steps, the most crucial part is the transition from 2D to 3D compressed pose data:
\begin{itemize}[label={\textcolor{white}{\textbullet}}]
\item \texttt{Step I:} First, we obtain the length of the skeleton through the 2D keypoint coordinates ($x$ and $y$), $a$ and $b$ represent indices that identify the two keypoints (or joints) forming a bone $L = \sqrt{(ax - bx)^2 + (ay - by)^2}$.
%
\item \texttt{Step II:} We compute the 3D rotation angles from 2D keypoints data: $A_{x}, A_{y}, A_{z} = \frac{\text{angle}_{x}, \text{angle}_{y}, \text{angle}_{z}}{\sqrt{\text{angle}_{x}^{2} + \text{angle}_{y}^{2} + \text{angle}_{z}^{2}}}$, where $A$ represents the normalized angles.
\item \texttt{Step III:} We define $P_x, P_y, P_z$ as the starting joint coordinates computed from 2D keypoints, and define $Q_x, Q_y, Q_z$ as the target coordinates in 3D space: $Q_x = P_x + L \times A_x$, $Q_y = P_y + L \times A_y$, $Q_z = P_z + L \times A_z$.
\end{itemize}
These mathematical formulas initialize the skeletal model by calculating skeletal length $L$, root node position, rotation angle, and 3D coordinates ($x$, $y$, $z$), serving as input for simulating 3D human skeletal motion. Compared to previous methods \cite{zelinka2020neural}, this approach is much more efficient, focusing solely on extracting the posture and gesture information relevant to sign language from the video data. This preprocessing step discards redundant information, reduces data size by 80\% relative to the raw video, and standardizes the format, enabling easier integration with text-to-text and sequence-to-sequence models and applications without the need for sign language-specific data loaders.

\noindent\textbf{Dataset Statistics.} \label{paragraph:prompt2sign_statistics}
After the data processing, train, dev, and test sets of different language parts are shown in Table~\ref{tab:dataset_statistics}.
We constructed 120 English templates and 210 prompt word templates for other languages (with 30 templates for each language), which are randomly associated with the oral text data to form the prompt part. The templates are carefully selected from hundreds of sentences generated by LLM that cover most everyday expressions. 




\section{Our Model: SignLLM} \label{sec:production_pipeline}

\subsection{Preliminary of Text2Pose Method} \label{subsec:text2pose}

The general SLP pipeline (\ie, text to sign language video) \cite{saunders2020progressive,saunders2021continuous,huang2021towards,saunders2021mixed,saunders2021skeletal} has following steps: text-to-gloss conversion, gloss-to-pose mapping, and finally pose-to-video rendering. 
In our work, we mainly focus on the first two steps.

\noindent\textbf{Text2Gloss \& Gloss2Pose.}\label{paragraph:text_to_pose}
Essentially, the transformation from text-to-gloss and gloss-to-pose can be distilled to a sequence-to-sequence \cite{saunders2020progressive,saunders2021continuous} problem in the realm of textual data, and their structures to bear significant resemblances.
We define $x_u$ 
as the input text $x$ tokens at position $u$ (total number is $\mathcal{U}$, position from 1 to $\mathcal{U}$), $p_{w}$ 
as the output pose $p$ at position $w$ (total number is $\mathcal{W}$, frame position from 1 to $W$), and then we use an encoder-decoder transformer framework to convert input into output:
\begin{equation}
\label{eq:T2P_encoder}
    f_{u} = Enc_{input2output}(x_{u} | x_{1:\mathcal{U}})
\end{equation}
\vspace{-12pt}
\begin{equation}
    p_{w+1} = Dec_{input2output}(p_{w}  | p_{1:w-1} , f_{1:\mathcal{U}})
\end{equation}
Here, $f_{u}$ denotes the encoded source of $x_u$. The output text tokens generated from this process form the input for the next stage of our translation model. In short, text2pose is the core method of SLP, and some researchers use gloss as an intermediate to make it text2gloss2pose.

\subsection{Design Overview} \label{subsec:Design_to_Overview}
\noindent\textbf{Architecture.} \label{paragraph:Basic_Idea} \ourLLMmethodName{} has two modes Multi-Language Switching Framework (MLSF) and Prompt2LangGloss, as shown in Fig. \ref{fig:Model_Overview} (Left), both make the model capable of multilingual sign language production by using the multilingual \ourDataName{} dataset. Both modes can be trained by new RL Loss and Priority Learning Channel module.

\noindent\textbf{Motivation.} 
Unlike existing SLP models, LLM directly fine-tuning for text-to-pose translation would hamper its dialogue capabilities, while using it without fine-tuning would treat translation requests as questions rather than performing actual text-to-pose translations. Therefore, we propose two specialized modes: MLSF for direct text-to-pose translation and Prompt2LangGloss for LLM-based interaction (\eg, \textit{``how to sign `thank you' in ASL?'' as input}). These modes, analogous to evolutionary branches of the same species, serve complementary purposes in multilingual SLP. 





%
\subsection{Two Multilingual SLP Modes} \label{subsec:MSLP}

MLSF is a mode that most existing models can refer to, like a dictionary/translator, and Prompt2LangGloss mode is designed specifically for sign language LLM. They are like two evolutionary branches diverged from the same species.

\noindent\textbf{Multi-Language Switching Framework.}
It can be understood as having multiple parallel Text2Pose channels/Enc-Dec groups, each language has an Enc-Dec group, allowing each channel/Enc-Dec to be independently trained and inferred.
Text2Pose visual representation is shown on the left of Fig. \ref{fig:Model_Overview}, the red rectangle represents the eight Enc-Dec in our model, and the middle partition represents the parameters stored separately in different Enc-Dec groups.
The assignment operation could be formalized as $\text{Enc}_{\mathcal{L}} = \mathcal{E}_\mathcal{L}$ and $\text{Dec}_{\mathcal{L}} = \mathcal{D}_\mathcal{L}$. Here, $\mathcal{L}$ denotes the language of input, while $\mathcal{E}_\mathcal{L}$ and $\mathcal{D}_\mathcal{L}$ are the mapping from language $\mathcal{L}$ to an encoder and decoder in the sets $\mathcal{E}$ and $\mathcal{D}$.
Similar to selecting tools from a drawer, MLSF allows you to choose the appropriate $\mathcal{E}_{ASL}$-$\mathcal{D}_{ASL}$ pair from the $\mathcal{E}$-$\mathcal{D}$ groups for training and inference. This modular design functions like a language drawer system, where each language Enc-Dec component can be accessed and utilized on demand from eight pairs.

\noindent\textbf{Prompt2LangGloss.}\label{sec:text_to_langgloss}
While MLSF tackles multilingual support through architectural design, Prompt2LangGloss approaches the challenge from a linguistic perspective. This mode introduces a novel intermediate representation that bridges the gap between natural language understanding and sign language generation. Enriching the traditional gloss notation with language-specific attributes creates a more nuanced and contextually aware translation process.

Gloss, essentially a shorter textual representation of sign language gestures, operates as an intermediate entity when using a text2pose model. As shown in Fig.~\ref{fig:Model_Overview} (Left), our proposed enhancement of this model involves appending an additional language attribute to each text word during the reading and tokenizing stages. For instance, a traditional gloss token ``\texttt{<xxx>}" can be transformed into ``\texttt{<ASL\_xxx>}", thus introducing a LangGloss layer of conditional input $f_{u} = Enc_{t2lg}(x_{u} | x_{1:\mathcal{U}})$ into \ac{slp} based on Eq.~\eqref{eq:T2P_encoder}:
$lg_{w+1} = Dec_{t2lg}(lg_{w}  | lg_{1:w-1} , f_{1:\mathcal{U}})$.
So, our LangGloss is a pseudo-Gloss used to distinguish language information in the parameter space by identifying language attributes.
This way, we solve several challenges: (1) LangGloss allows the existing models to train multilingual data. 
By adding language attributes to gloss, we reduce the semantic ambiguity that occurs when words share the same form across different languages.
(2) LangGloss, as a mediator, can solve the limitations of the existing model in understanding complex, natural human inputs. It reduces the negative impact of directly processing intricate prompt words, improving the model's response accuracy.

In short, these two modes create a robust multilingual SLP system: 
This dual-mode design enables \ourLLMmethodName{} to achieve processing efficiency and semantic accuracy in existing models and sign language LLMs, respectively.





\subsection{Reinforcement Learning Training Strategy} \label{sec:RL_components}

\noindent To reduce training time, we utilize the RL reward concept to quantify each training batch's quality and prioritize valuable batches through the Priority Learning Channel module. Both Multilingual SLP modes can use RL Loss and PLC.

\noindent\textbf{Reinforcement Learning Loss.}
Reinforcement Learning (RL) has an important advantage in identifying high-value actions or samples, allowing for prioritized learning of valuable data. This approach could help address the challenge of slow training when using more sign languages. But before learning high-value data batch, we should transform the ordinary generative model first into an RL-like model, so we design RL Loss for model transformation.

\begin{figure}[ht]
    \centering
    \includegraphics[width=\linewidth]{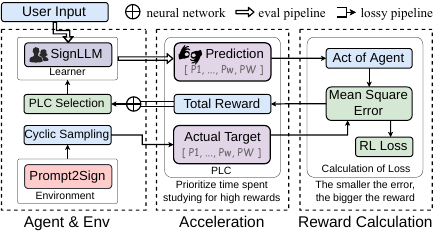}
    \vspace{-22pt}
    \caption{RL elements: \textit{User, Agent, Environment, Cyclic Sampling, PLC} to sketch the sequence prediction learning process.}
    \label{fig:RL_Loss}
    \vspace{-14pt}
\end{figure}%

Concretely, we set the input sequence as the state $s_t$, and the output sequence is the action $a_t$, and the reward $r_t$. $t$ stands for time, $i$ stands sample, $\hat{y}_i$ represents the model's predicted output, and $N$ denotes the total number of samples. The closer the prediction is to reality (mean squared error), the greater the reward: $r = -\frac{1}{N}\sum_{i=1}^{N}(y_i-\hat{y}_i)^2$.
With this interpretation, we can reformulate the traditional supervised learning problem of minimizing MSE loss to maximize the expected cumulative reward, where $L$ denotes the MSE loss function, $M$ is the model, and $x_t$, $y_t$ are the model inputs and corresponding targets respectively:
\begin{equation}
\footnotesize
\begin{aligned}
\theta^* = \underset{\theta}{\operatorname{argmax}} \, E_{\theta}\left[\sum_{t=0}^{T} r_t\right]  = \underset{\theta}{\operatorname{argmin}} \, E_{\theta}\left[\sum_{t=0}^{T} L(y_t, M(x_t))\right]
\end{aligned}
\end{equation}
where $\theta$ represents the trainable parameters of the model, $E_{\theta}$ denotes the expected value with respect to the model parameters $\theta$, and $T$ is the total number of time steps in the sequence.
Here, $\operatorname{argmax}$/$\operatorname{argmin}$ returns the value of $\theta$ that maximizes/minimizes the expected cumulative reward.
These optimized parameters $\theta$ are founded by using gradient descent, updating parameters proportionally to the gradient of expected cumulative reward concerning model parameters. 
In summary, we quantify the effectiveness of MSE optimization at each time step as reward $r$. Then the $r$ will serve as the input for our Priority Learning Channel.

\noindent\textbf{Priority Learning Channel.} 
The RL Loss itself does not possess subjective acceleration capabilities, it is designed for PLC to prioritize the learning of more valuable data.
We have defined rewards $r$, sample $i$, and data for each batch $j$. They then are converted into sampling probabilities for each data sample according to $P(i) = \frac{ r(i)^\eta}{\sum_{j \in S} r(j)^\eta}$, 
where $\eta$ regulates the intensity of prioritization, and $S$ represents the dataset. By employing these sampling probabilities, the choice of data samples for each batch is no longer uniform but regulated by their respective rewards (\eg, if the reward is less than 50\%, skip the batch). 
The per-step RL loss, $L(i)$, is computed for the chosen data, which is then used to optimize the model parameters following the policy gradient theorem. This procedure is formally expressed as $Minimize \; \; E_{i \sim P(i)}[L(i)]$ (where $E$ is the expectation), the whole RL Loss and PLC system is shown in Fig. \ref{fig:RL_Loss}.
By continually updating the model based on the most rewarding samples, the PLC brings the advantages of RL to 
sequence prediction tasks. The adaptive nature of the PLC ensures that the model's focus shifts by the model's evolving knowledge, thereby accelerating the learning process.


\begin{table*}[t]
\centering
\resizebox{1\linewidth}{!}{%
\begin{tabular}{@{}p{2.8cm}ccccc|ccccc@{}}
\toprule
\midrule
 & \multicolumn{5}{c}{DEV SET} & \multicolumn{5}{c}{TEST SET} \\ 
\multicolumn{1}{c|}{Type:} & BLEU-4         & BLEU-3         & BLEU-2         & BLEU-1         & ROUGE          & BLEU-4         & BLEU-3         & BLEU-2         & BLEU-1         & ROUGE          \\ \midrule
\multicolumn{1}{r|}{NSLP-G \cite{hwang2021non}}   & - & - & - & - & - & 5.75 & 8.21 & 11.62 & 17.55 & 31.98 \\
\multicolumn{1}{r|}{Fast-SLP Transformers \cite{fang2024signdiffdiffusionmodelsamerican}}  & 17.19 & 23.11 & 29.49 & 36.96 & 55.85 & 12.85 & 17.35 & 23.38 & 39.46 & 46.89 \\
\multicolumn{1}{r|}{Neural Sign Actors \cite{Baltatzis_2024_CVPR}}   & - & - & - & - & - & 13.12 & 18.25 & 25.44 & 41.31 & 47.55 \\
\midrule
\multicolumn{1}{r|}{\textbf{SignLLM-1x40M-Base-M (ASL) }}   & 18.77 & 25.42 & 32.44 & 40.66 & 61.44 & 14.13 & 19.08 & 25.72 & 43.40 & 51.57 \\
\multicolumn{1}{r|}{\textbf{SignLLM-1x120M-Large-M (ASL) }}   & 19.40 & 26.11 & 33.38 & 41.89 & 63.21 & 14.52 & 19.65 & 26.44 & 44.72 & 53.14 \\
\multicolumn{1}{r|}{\textbf{SignLLM-1x1B-Super-M (ASL) }}   & \textbf{\quad 20.09}\textcolor{sgreen}{\small{\textbf{ +2.90}}} & \textbf{27.04}\textcolor{sgreen}{\small{\textbf{ +3.93}}} & \textbf{34.45}\textcolor{sgreen}{\small{\textbf{ +4.96}}} & \textbf{43.16}\textcolor{sgreen}{\small{\textbf{ +6.20}}} & \textbf{65.19}\textcolor{sgreen}{\small{\textbf{ +9.36}}} & \textbf{\quad 15.03}\textcolor{sgreen}{\small{\textbf{ +2.18}}} & \textbf{20.28}\textcolor{sgreen}{\small{\textbf{ +2.93}}} & \textbf{27.35}\textcolor{sgreen}{\small{\textbf{ +3.97}}} & \textbf{46.17}\textcolor{sgreen}{\small{\textbf{ +6.71}}} & \textbf{54.86}\textcolor{sgreen}{\small{\textbf{ +7.97}}} \\
\midrule
\multicolumn{1}{r|}{\textbf{\color{cancelgray} SignLLM-1x40M-Base-P (ASL) }}   & \color{cancelgray} 17.34 & \color{cancelgray} 23.57 & \color{cancelgray} 29.87 & \color{cancelgray} 37.81 & \color{cancelgray} 56.93 & \color{cancelgray} 13.06 & \color{cancelgray} 17.66 & \color{cancelgray} 23.77 & \color{cancelgray} 40.15 & \color{cancelgray} 47.76 \\
\multicolumn{1}{r|}{\textbf{\color{cancelgray} SignLLM-1x120M-Large-P (ASL) }}   & \color{cancelgray} 18.05 & \color{cancelgray} 24.28 & \color{cancelgray} 31.04 & \color{cancelgray} 38.97 & \color{cancelgray} 58.78 & \color{cancelgray} 13.48 & \color{cancelgray} 18.27 & \color{cancelgray} 24.57 & \color{cancelgray} 41.57 & \color{cancelgray} 49.42 \\
\multicolumn{1}{r|}{\textbf{\color{cancelgray} SignLLM-1x1B-Super-P (ASL) }}   & \color{cancelgray} \quad 18.68\textcolor{sgreen}{\small{\textbf{ +1.49}}} & \color{cancelgray} 25.11\textcolor{sgreen}{\small{\textbf{ +2.00}}} & \color{cancelgray} 31.99\textcolor{sgreen}{\small{\textbf{ +2.50}}} & \color{cancelgray} 40.14\textcolor{sgreen}{\small{\textbf{ +7.18}}} & \color{cancelgray} 60.47\textcolor{sgreen}{\small{\textbf{ +4.62}}} & \color{cancelgray} \quad 13.93\textcolor{sgreen}{\small{\textbf{ +0.92}}} & \color{cancelgray} 18.86\textcolor{sgreen}{\small{\textbf{ +1.51}}} & \color{cancelgray} 25.40\textcolor{sgreen}{\small{\textbf{ +2.02}}} & \color{cancelgray} 42.87\textcolor{sgreen}{\small{\textbf{ +3.41}}} & \color{cancelgray} 50.91\textcolor{sgreen}{\small{\textbf{ +4.02}}} \\
\midrule
\bottomrule
\end{tabular}%
}
\vspace{-5pt}
\caption{
\textbf{\acf{aslp}:} Comparison of \ourLLMmethodName{} variants with baseline on \textit{Text to Pose} task by using our \ourDataName{} ASL part. 
``-'' : The NSA \cite{Baltatzis_2024_CVPR} and NSLP-G \cite{hwang2021non} have not been tested on the dev set, and there is no source code.
The improvement (\textcolor{sgreen}{\small{\textbf{+num}}}) is relative to the latest work \cite{fang2024signdiffdiffusionmodelsamerican}.
}\label{tab:aslp}
\vspace{4pt}
\end{table*}

\begin{table*}[!ht]
\vspace{-6pt}
\centering
\resizebox{1\linewidth}{!}{%
\begin{tabular}{@{}p{2.8cm}c|ccccc|ccccc@{}}
\toprule
\midrule
  & & \multicolumn{5}{c}{DEV SET} & \multicolumn{5}{c}{TEST SET} \\ 
\multicolumn{1}{c|}{Type:} & \multicolumn{1}{c|}{Language:} & BLEU-4         & BLEU-3         & BLEU-2         & BLEU-1         & ROUGE          & BLEU-4         & BLEU-3         & BLEU-2         & BLEU-1         & ROUGE          \\ \midrule
\multicolumn{1}{r|}{\textbf{SignLLM-6x40M-Base-M}} & DSGS  & 9.73 & 15.82 & 19.85 & 24.84 & 37.57 & 7.34 & 9.89 & 16.85 & 26.04 & 31.51 \\
\multicolumn{1}{r|}{\textbf{SignLLM-6x120M-Large-M}} & DSGS   & \textbf{11.45} & 17.28 & 19.84 & 29.07 & \textbf{41.69} & \textbf{9.69} & 14.06 & 16.35 & 29.80 & \textbf{31.51} \\
\midrule
\multicolumn{1}{r|}{\textbf{SignLLM-6x40M-Base-M}} & LSF-CH  & 9.79 & 18.48 & 23.13 & 28.86 & 34.98 & 8.92 & 13.12 & 16.11 & 26.48 & 37.92 \\
\multicolumn{1}{r|}{\textbf{SignLLM-6x120M-Large-M}} &  LSF-CH   & \textbf{13.72} & 20.79 & 23.40 & 25.15 & \textbf{38.39} & \textbf{9.60} & 12.58 & 16.98 & 22.71 & \textbf{41.96} \\
\midrule
\multicolumn{1}{r|}{\textbf{SignLLM-6x40M-Base-M}} & LIS-CH  & 10.81 & 14.46 & 19.93 & 24.55 & 35.83 & 7.34 & 10.56 & 15.24 & 22.73 & 36.42 \\
\multicolumn{1}{r|}{\textbf{SignLLM-6x120M-Large-M}} & LIS-CH   & \textbf{12.10} & 18.04 & 23.01 & 25.95 & \textbf{36.98} & \textbf{9.30} & 11.20 & 15.68 & 23.38 & \textbf{38.37} \\
\midrule
\multicolumn{1}{r|}{\textbf{SignLLM-6x40M-Base-M}} & LSA  & 10.72 & 15.55 & 21.76 & 25.91 & 38.78 & 7.33 & 14.86 & 16.68 & 22.55 & 34.42 \\
\multicolumn{1}{r|}{\textbf{SignLLM-6x120M-Large-M}} & LSA  & \textbf{11.69} & 14.79 & 26.25 & 28.08 & \textbf{39.01} & \textbf{8.21} & 11.04 & 17.05 & 26.68 & \textbf{37.46} \\
\midrule
%
\multicolumn{1}{r|}{\textbf{SignLLM-6x40M-Base-M}} & KSL & 9.42 & 14.67 & 17.24 & 26.41 & 31.96 & 8.31 & 11.84 & 17.93 & 24.15 & \textbf{33.78} \\
\multicolumn{1}{r|}{\textbf{SignLLM-6x120M-Large-M}} & KSL & \textbf{12.91} & 19.45 & 15.17 & 24.07 & \textbf{37.83} & \textbf{10.09} & 13.06 & 18.37 & 25.75 & 33.69 \\
\midrule
\multicolumn{1}{r|}{Hybrid Translation System \cite{8778347}} & TSL  & - & - & - & - & - & 12.64 & 18.28 & 31.48 & 53.17 & - \\
\multicolumn{1}{r|}{\textbf{SignLLM-6x40M-Base-M}}  & TSL & 14.53 & 19.86 & 29.93 & 36.86 & 58.01 & 13.23 & 17.80 & 25.39 & 39.30 & 57.03 \\
\multicolumn{1}{r|}{\textbf{SignLLM-6x120M-Large-M}}  & TSL & \textbf{15.17} & 21.70 & 31.73 & 38.86 & \textbf{71.10} & \textbf{14.36} & 18.74 & 26.96 & 43.21 & \textbf{57.12}\\
\midrule
\bottomrule
\end{tabular}%
}
\vspace{-5pt}
\caption{
\textbf{\acf{mslp}:} Comparison of different \ourLLMmethodName{} M-mode variants with a baseline on \textit{Text to Pose} task. We propose the first Multilingual SLP benchmark, with the exception of the existing TSL-Baseline.
}\label{tab:mslp}
\vspace{-6pt}
\end{table*}

\section{Experiments and Discussions} \label{sec:experiments}

\textbf{Setup.}
We provide the naming rules as follows: SignLLM-\{\textcolor[rgb]{0.756,0.121,0.121}{number of languages}\}x\{\textcolor{cvprblue}{single language parameters}\}-\{\textcolor[rgb]{1,0.576,0.098}{submode size}\}-\{\textcolor{violet}{the mode of training}\}-\{the language of input\}. Such as ``\ourLLMmethodName{}-\textcolor[rgb]{0.756,0.121,0.121}{2}x\textcolor{cvprblue}{40M}-\textcolor[rgb]{1,0.576,0.098}{Base}-\textcolor{violet}{M} (ASL)'', the nomenclature ``2x40'' denotes that the model comprises 2 language knowledge, with each language component estimated to be around 40 million parameters in size, and a total is 80 million parameters (``1B'' represents a total of 1 billion parameters). There are Base, Large, and Super versions, depending on a single language parameter size provided by the model. 
The encoder and decoder of our model versions (\ie, Base, Large, Super) both have two layers. When the model is expanded to Large and Super versions, the layers are unchanged, the parameters are expanded by about two and four times, respectively. 
M and P stand for models trained using MLSF and Prompt2LangGloss. At the end is the language of the input model, 
ASL, GSL, LSA \etc. 




\begin{table}[ht]
\vspace{6pt}
\centering
\resizebox{0.99\linewidth}{!}{%
\begin{tabular}{@{}p{2.8cm}cc|cc@{}}
\toprule
\midrule
     & \multicolumn{2}{c}{DEV SET}  & \multicolumn{2}{c}{TEST SET} \\ 
\multicolumn{1}{c}{Approach:}  & BLEU-4  & ROUGE & BLEU-4 & ROUGE \\ \midrule
\multicolumn{1}{r|}{Progressive Transformers \cite{saunders2020progressive}} & 11.82 & 33.18 & 10.51 & 32.46 \\ 
\multicolumn{1}{r|}{Adversarial Training \cite{saunders2020adversarial}} & 12.65 & 33.68 & 10.81 & 32.74 \\
\multicolumn{1}{r|}{Mixture Density Networks \cite{saunders2021continuous}} & 11.54 & 33.40 & 11.68 & 33.19 \\ 
\multicolumn{1}{r|}{Mixture of Motion Primitives \cite{saunders2021mixed}} & 14.03 & 37.76 & 13.30 & 36.77 \\
\multicolumn{1}{r|}{Photo-realistic SLP \cite{Saunders_2022_CVPR}}  & 16.92  & 35.74 & 21.10 & 42.57 \\
\multicolumn{1}{r|}{Fast-SLP Transformers \cite{fang2024signdiffdiffusionmodelsamerican}} & 18.26  & 39.62 & 22.15 & 46.82 \\
\midrule
\multicolumn{1}{r|}{\textbf{SignLLM-1x40M-Base-M (GSL)}} & 18.61  & 40.69 & 22.76 & 48.05 \\
\multicolumn{1}{r|}{\textbf{SignLLM-1x120M-Large-M (GSL)}} & \textbf{\quad 19.31}\textcolor{sgreen}{\small{\textbf{ +1.05}}}  & 41.42 & \textbf{\quad 23.25}\textcolor{sgreen}{\small{\textbf{ +1.10}}} & 49.08 \\
\multicolumn{1}{r|}{\textbf{SignLLM-1x1B-Super-M (GSL)}} & 19.07  & \textbf{41.83}\textcolor{sgreen}{\small{\textbf{ +2.21}}} & 23.21 & \textbf{49.52}\textcolor{sgreen}{\small{\textbf{ +2.70}}} \\
\midrule
\multicolumn{1}{r|}{\textbf{\color{cancelgray} SignLLM-1x40M-Base-P (GSL)}} & \color{cancelgray} 17.12  & \color{cancelgray} 37.43 & \color{cancelgray} 20.93 & \color{cancelgray} 44.21 \\
\multicolumn{1}{r|}{\textbf{\color{cancelgray} SignLLM-1x120M-Large-P (GSL)}} & \color{cancelgray} 17.55  & \color{cancelgray} 38.10 & \color{cancelgray} 21.39 & \color{cancelgray} 45.16 \\
\multicolumn{1}{r|}{\textbf{\color{cancelgray} SignLLM-1x1B-Super-P (GSL)}} & \color{cancelgray} 17.54  & \color{cancelgray} 38.48 & \color{cancelgray} 21.35 & \color{cancelgray} 45.57 \\
\midrule
\bottomrule
\end{tabular}%
}
\caption{\textbf{German \acf{slp}:} Comparison of different models with existing work on \textit{Text to Pose} task. 
The improvement (\textcolor{sgreen}{\small{\textbf{+num}}}) is relative to the latest work \cite{fang2024signdiffdiffusionmodelsamerican}.
}\label{tab:gslp}
\vspace{-0.7cm}
\end{table}

\noindent\textbf{Metrics.}
\texttt{(i) BLEU-n score} measures the similarity between machine-generated translations and reference translations based on n-grams, the closer the predicted result is to the input (reference), the higher the value. BLEU-n \cite{bleu} means that n words are used as the basic computing unit, and the higher the n, the higher the fluency requirement.
\texttt{(ii) ROUGE score} \cite{lin-acl-2004-rouge} is similar to BLEU, but it is more concerned with consistency and coverage. It indicates better agreement between the generated and reference texts, indicating a more accurate and comprehensive summary.
\texttt{(iii) DTW score} underpinned by dynamic programming principles \cite{berndt1994dtw}, is employed to ascertain the smallest manipulation distance between clips and sentences; the lower, the better. 



\begin{table*}[ht]
\centering
\resizebox{1\linewidth}{!}{%
\begin{tabular}{@{}p{2.8cm}ccccc|ccccc@{}}
\toprule
\midrule
     & \multicolumn{5}{c}{DEV SET}  & \multicolumn{5}{c}{TEST SET} \\ 
\multicolumn{1}{c|}{Approach:}  & BLEU-4         & BLEU-3         & BLEU-2         & BLEU-1         & ROUGE          & BLEU-4         & BLEU-3         & BLEU-2         & BLEU-1         & ROUGE          \\ \midrule
\multicolumn{1}{r|}{\textbf{\color{cancelgray} Base + Normal MSE Loss}} & \color{cancelgray} 10.96 & \color{cancelgray} 14.68 & \color{cancelgray} 22.49 & \color{cancelgray} 40.29 & \color{cancelgray} 44.13 & \color{cancelgray} 9.27 & \color{cancelgray} 10.72 & \color{cancelgray} 18.64 & \color{cancelgray} 39.88 & \color{cancelgray} 40.39 \\
\multicolumn{1}{r|}{\textbf{Base + RL Loss}}    & 18.33 & 24.28 & 32.18 & 48.83 & 52.61 & 13.26 & 17.33 & 24.51 & 40.76 & 40.95 \\ 
\multicolumn{1}{r|}{\textbf{Base + RL Loss \& PLC}}       & \textbf{18.77} & \textbf{25.42} & \textbf{32.44} & \textbf{52.66} & \textbf{61.44} & \underline{14.13} & \underline{19.08} & \underline{25.72} & \underline{43.40} & \textbf{51.57} \\
\multicolumn{1}{r|}{\textbf{Base + MSE + Prompt2LangGloss}}& 15.78 & 22.66 & 30.27 & 48.30 & 50.88 & 16.32 & 20.27 & 28.70 & 47.89 & 48.18\\ 
\multicolumn{1}{r|}{\textbf{Base + MSE + MLSF}} & 16.44 & 23.79 & 32.32 & 50.97 & 53.44 & \textbf{17.17} & \textbf{21.25} & \textbf{30.12} & \textbf{50.25} & 50.47 \\
\midrule
\bottomrule
\end{tabular}%
}
\vspace{-6pt}
\caption{\textbf{Comparison of Different Modules:} Base: \ourLLMmethodName{}-40M-Base results for \textit{Text to Pose} task on the ASL part of \ourDataName{}.
PLC: Priority Learning Channel. 
Top performances are highlighted in \textbf{bold}, while second top performances are \underline{underlined}.}\label{tab:data_augmentation_results}
\vspace{-12pt}
\end{table*}

\break

\subsection{Quantitative Evaluation} \label{sec:ASL_experiments}

\textbf{Back Translation.}
Back translation means translating generated sign language videos back into spoken language sentences. These sentences are compared with the original input sentences to evaluate the translation quality. The task is widely adopted to evaluate \acf{slp} as it can indicate the accuracy of the produced sign language videos \cite{saunders2020progressive}, our translation models \cite{camgoz2020sign} are trained on the corresponding language Prompt2Sign pose data, the performance is between bleu-4 22.3 and 27.6, it is comparable to the translation model used in previous work.

In Table \ref{tab:aslp}, we conduct American Sign Language Production back-translation tests using \ourLLMmethodName{} on the ASL part of our new dataset, and Table \ref{tab:gslp} further compares our method with other recent approaches for German SLP on the GSL part of \ourDataName{} dataset \cite{saunders2020progressive,saunders2020adversarial,saunders2021continuous,saunders2021mixed,fang2024signdiffdiffusionmodelsamerican}. 
These two languages stand for high-resource languages (\ie, languages with rich data resources), and our comprehensive tests at different levels on the dataset demonstrate impressive performance compared with the latest works in the field.
The results affirm the competitiveness and potential of our proposed method, regardless of the specific sign language in use. More evaluation results and analyses for these two languages can be found in the supplementary materials.

In Table \ref{tab:mslp}, we present the results of our model for six different sign languages. These six languages represent low-resource languages, and they are considered limited languages. For languages with low resources, their vocabulary, video time, and diverse corpus sources are relatively low, making training more difficult.
From the table data, it can be observed that our performance remains strong in languages where training data is lacking. As long as the input text/prompt can be encoded as a computationally recognizable word and video exists, our method can translate it into the corresponding language pose video after training.




\subsection{Ablation Evaluation}

\begin{figure}[t]
    \centering
    \includegraphics[width=0.99\linewidth]{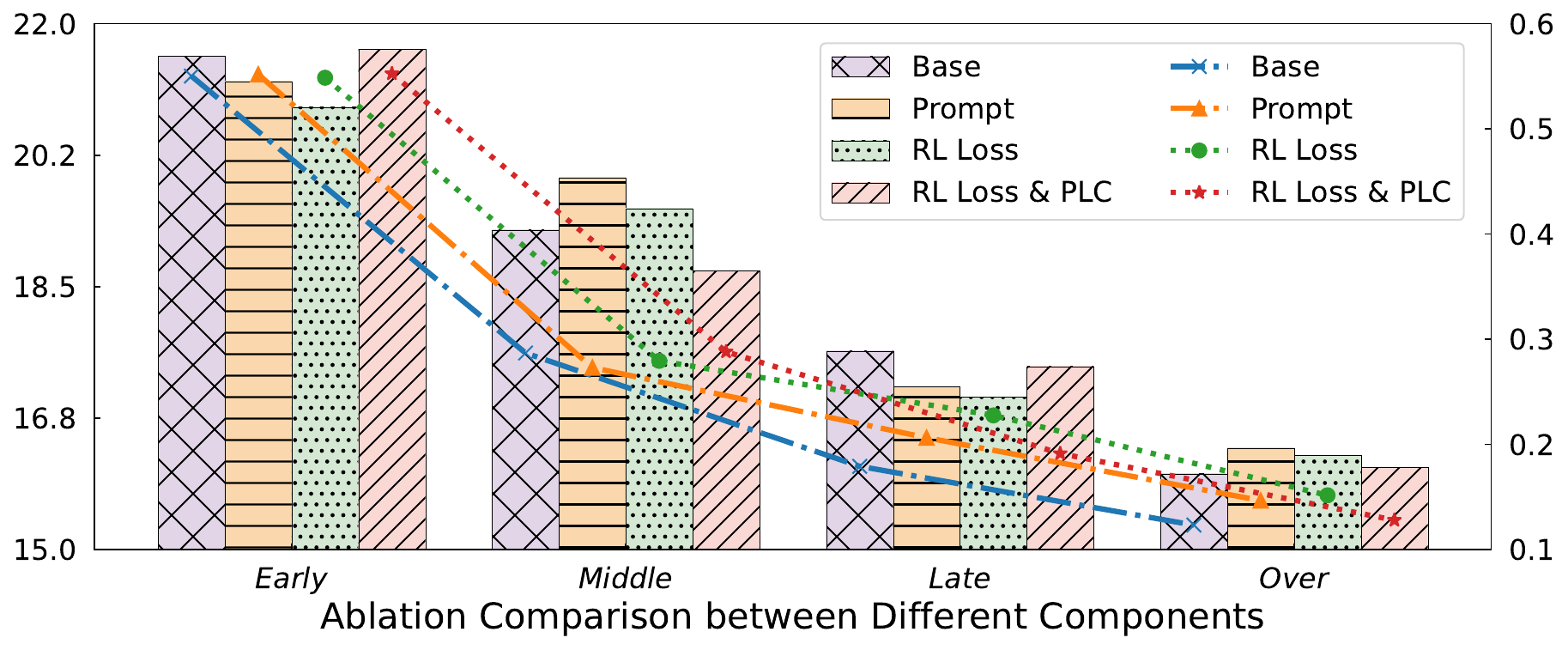}
    \vspace{-4pt}
    \caption{\textbf{RL Training Efficiency Analysis:} Comparison of different Settings on DTW values (the lower the better) at different peroid (every 30 epochs a peroid). Left Y-axis: Value of DTW. Right Y-axis: Value of Loss. Prompt: Prompt2LangGloss mode.}
    \label{fig:former_barline}
\vspace{-12pt}
\end{figure}%

\noindent\textbf{Performance.} Ablation results in Table \ref{tab:data_augmentation_results} indicates our four innovative strategies (Prompt2LangGloss, MLSF modes, RL Loss and PLC module) significantly improve the model's performance, and four strategies contribute to substantial improvements. 
When replacing the standard MSE Loss with RL Loss, we observe a significant improvement in the model’s performance. These results show that PLC has successfully enhanced the adaptability of the original RL Loss to unknown environments. Incorporating the PLC module further enhances these gains. We also note that the base model with MSE, when upgraded with Prompt2LangGloss, achieves higher scores with the new mode, slightly underperforming compared to the MLSF mode but far surpassing the baseline model in the first row. This indicates that, with sufficient data, the Prompt2LangGloss mode demonstrates good usability, and we explore this further in the supplementary materials.


In Fig.~\ref{fig:former_barline}, we compare the Base model's performance with different module schemes. The ablation comparison is primarily based on observing the variability of DTW scores across epochs to assess the effectiveness of each approach. We observe that: (i) The standalone use of the Prompt2LangGloss model exhibits the lowest efficiency, as it introduces noise by incorporating prompts and tokenizers, which is also evident in Table \ref{tab:data_augmentation_results}. (ii) The combination of the two RL methods shows modest performance improvement compared to using each method individually, although it is not statistically significant. Future work could explore adaptive weighting mechanisms or hierarchical integration of these methods to leverage their complementary strengths fully. (iii) In terms of training efficiency, they significantly reduce training time. Compared to the base setting, our RL Loss with PLC approach reduces training time by 27.1\%, which is crucial for training large-scale sign language models.
Our ablation studies reveal that RL-based sample prioritization not only enhances training efficiency but also provides a scalable, unified, multilingual SLP framework.






\begin{figure*}[t!]
    \centering
    \includegraphics[width=1\linewidth]{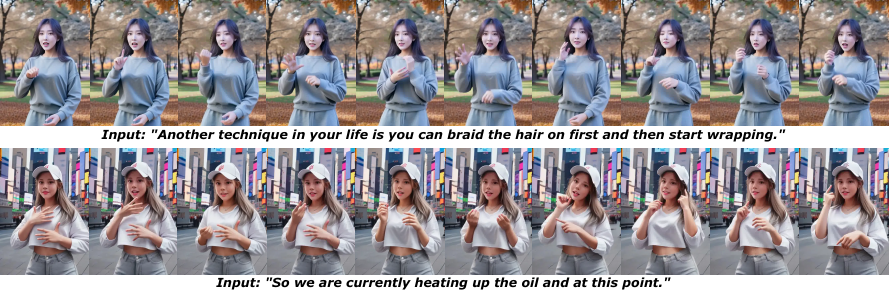}
    \vspace{-22pt}
    \caption{We use an adjusted vid2vid model \cite{feng2023dreamoving} to convert the predicted skeletal pose video into a more realistic final video.
    }
    \label{fig:QP}
    \vspace{-12pt}
\end{figure*}%

\break

\subsection{Qualitative Evaluation} \label{sec:qualitative_experiments}

\noindent\textbf{Qualitative Presentation.}
We use our predicted pose results as input, which are then used to generate rendered videos in Fig. \ref{fig:QP}. We can observe that our video outcomes are of high quality, with highly accurate finger movements and high image fidelity. Our results surpass all previous works, benefiting not only from technological advancements but also from the superior output quality of our \ourLLMmethodName{} compared to previous smaller models: the postures we predict are rarely missing, unlike previous works that often suffered from issues such as flickering, incomplete or missing fingers, and low input quality due to densely packed fingers, so it works well with the latest model \cite{feng2023dreamoving}. However, the final sign language video finger problem still exists anyway, and in the future, we can consider the special optimization of these style transfer vid2vid models.

\begin{table}[ht]
\vspace{-0.40cm}
\centering
\resizebox{0.99\linewidth}{!}{%
\begin{tabular}{@{}p{2.8cm}cc|cc@{}}
\toprule
\midrule
     & \multicolumn{2}{c}{DEV SET}  & \multicolumn{2}{c}{TEST SET} \\ 
\multicolumn{1}{c}{Approach:}  & BLEU-4  & ROUGE & BLEU-4 & ROUGE \\ \midrule
\multicolumn{1}{r|}{Progressive Transformers \cite{saunders2020progressive}} & 10.79 & 36.15 & 9.59 & 35.42 \\ 
\multicolumn{1}{r|}{Fast-SLP Transformers \cite{fang2024signdiffdiffusionmodelsamerican}} & 16.68 & 43.2 & 24.24 & 51.05 \\
\midrule
\multicolumn{1}{r|}{\textbf{SignLLM-1x40M-Base-M (GSL)}} & 16.96 & 44.41 & 24.74 & 52.45 \\
\multicolumn{1}{r|}{\textbf{SignLLM-1x120M-Large-M (GSL)}} & \textbf{\quad 17.73}\textcolor{sgreen}{\small{\textbf{ +1.05}}} & 45.11 & \textbf{\quad 25.39}\textcolor{sgreen}{\small{\textbf{ +1.15}}} & 53.45 \\
\midrule
\multicolumn{1}{r|}{\textbf{\color{cancelgray} SignLLM-1x40M-Base-P (GSL)}} & \color{cancelgray} 16.27 & \color{cancelgray} 43.76 & \color{cancelgray} 24.12 & \color{cancelgray} \textbf{54.35}\textcolor{sgreen}{\small{\textbf{ +3.30}}} \\
\multicolumn{1}{r|}{\textbf{\color{cancelgray} SignLLM-1x120M-Large-P (GSL)}}& \color{cancelgray} 16.71 & \color{cancelgray} \textbf{45.40}\textcolor{sgreen}{\small{\textbf{ +2.20}}} & \color{cancelgray} 25.04 & \color{cancelgray} 52.80 \\
\midrule
\bottomrule
\end{tabular}%
}
\vspace{-4pt}
\caption{Presentation Effect Study: Results of \textit{Text to Sign} task in GSL. The improvement (\textcolor{sgreen}{\small{\textbf{+num}}}) is relative to the latest work \cite{fang2024signdiffdiffusionmodelsamerican}.
}\label{tab:text2sign}
\vspace{-10pt}
\end{table}

In Table \ref{tab:text2sign}, we conducted a series of final video back-translation evaluations (as shown in Fig. \ref{fig:QP}) based on the German SLP task to investigate two main research questions: (i) Whether rendering the predicted results into real sign language videos would lead to a decrease in accuracy. (ii) How does our work compare to previous studies on the task of text-to-sign-language real video generation? Based on our observations, there was generally not a significant accuracy loss, but there were some fluctuations compared to the baseline. Our approach outperformed previous works, which could be attributed to the higher quality of our data, making it more suitable as input for style transfer models.

\subsection{Discussion} \label{sec:discussion}

\textbf{Societal Impact.}
Our model has the potential to assist people with disabilities in three key areas: sign language teaching, generative sign language translation, and real-time interpretation for broadcasting. (1) Traditional sign language teaching relies heavily on human instructors and static pictorial representations, limiting learning accessibility. (2) Current sign language translation software remains inadequate for effective communication between deaf people and their family members who lack sign language knowledge. (3) Additionally, real-time sign language interpretation is only available for limited content like major news broadcasts, creating barriers for the deaf community in daily life. However, its current level of accuracy is not high enough to be fully trusted, and users must be cautious to use it.

\noindent\textbf{Limitation.} Our tool enhances sign language data processing automation but isn't fully end-to-end. Manual preprocessing is still required for OpenPose processing, video editing, and transcript alignment, \etc. For pose video to final video conversion, when we use a style transfer model to make the final video, it requires more complex processing of the pose video. So there is still a way to go before large-scale use, which requires us to make industrial-level adaptation improvements to the output style of \ourLLMmethodName{}.



\section{Conclusion} \label{sec:conclusion}

We present \ourLLMmethodName{}, a large multilingual SLP model. For training this model, we propose \ourDataName{}, a standardized dataset that contains eight sign languages.
Our model with two modes, MLSF and Prompt2LangGloss, progressively incorporates more sign languages while preserving translation efficiency and introducing LLM capabilities.
Our new RL loss and new PLC module solve the challenge of longer training time due to more data.
Finally, we show baseline comparisons, ablation studies, experiments under various parameters, and qualitative evaluations for discussion, which proves the efficacy of our methodology.

{
    \small
    \bibliographystyle{ieeenat_fullname}
    \bibliography{ref/main, ref/sds, ref/meta, ref/llm, ref/rl, ref/slt, ref/how2sign}

\begin{thebibliography}{111}
\providecommand{\natexlab}[1]{#1}
\providecommand{\url}[1]{\texttt{#1}}
\expandafter\ifx\csname urlstyle\endcsname\relax
  \providecommand{\doi}[1]{doi: #1}\else
  \providecommand{\doi}{doi: \begingroup \urlstyle{rm}\Url}\fi

\bibitem[Balakrishnan et~al.(2018)Balakrishnan, Zhao, Dalca, Durand, and Guttag]{balakrishnan2018synthesizing}
Guha Balakrishnan, Amy Zhao, Adrian~V Dalca, Fredo Durand, and John Guttag.
\newblock {Synthesizing Images of Humans in Unseen Poses}.
\newblock In \emph{Proceedings of the IEEE Conference on Computer Vision and Pattern Recognition (CVPR)}, 2018.

\bibitem[Baltatzis et~al.(2024)Baltatzis, Potamias, Ververas, Sun, Deng, and Zafeiriou]{Baltatzis_2024_CVPR}
Vasileios Baltatzis, Rolandos~Alexandros Potamias, Evangelos Ververas, Guanxiong Sun, Jiankang Deng, and Stefanos Zafeiriou.
\newblock Neural sign actors: A diffusion model for 3d sign language production from text.
\newblock In \emph{Proceedings of the IEEE/CVF Conference on Computer Vision and Pattern Recognition (CVPR)}, 2024.

\bibitem[Berndt and Clifford(1994)]{berndt1994dtw}
Donald~J Berndt and James Clifford.
\newblock {Using Dynamic Time Warping to Find Patterns in Time Series}.
\newblock In \emph{AAA1-94 Workshop on Knowledge Discovery in Databases}, 1994.

\bibitem[Boh\'a\v{c}ek and Hr\'uz(2022)]{Bohacek_2022_WACV}
Maty\'a\v{s} Boh\'a\v{c}ek and Marek Hr\'uz.
\newblock Sign pose-based transformer for word-level sign language recognition.
\newblock In \emph{Proceedings of the IEEE/CVF Winter Conference on Applications of Computer Vision (WACV) Workshops}, pages 182--191, 2022.

\bibitem[Brown et~al.(2020)Brown, Mann, Ryder, Subbiah, Kaplan, Dhariwal, Neelakantan, Shyam, Sastry, Askell, Agarwal, Herbert{-}Voss, Krueger, Henighan, Child, Ramesh, Ziegler, Wu, Winter, Hesse, Chen, Sigler, Litwin, Gray, Chess, Clark, Berner, McCandlish, Radford, Sutskever, and Amodei]{Brown-NeurIPS-2020-Language}
Tom~B. Brown, Benjamin Mann, Nick Ryder, Melanie Subbiah, Jared Kaplan, Prafulla Dhariwal, Arvind Neelakantan, Pranav Shyam, Girish Sastry, Amanda Askell, Sandhini Agarwal, Ariel Herbert{-}Voss, Gretchen Krueger, Tom Henighan, Rewon Child, Aditya Ramesh, Daniel~M. Ziegler, Jeffrey Wu, Clemens Winter, Christopher Hesse, Mark Chen, Eric Sigler, Mateusz Litwin, Scott Gray, Benjamin Chess, Jack Clark, Christopher Berner, Sam McCandlish, Alec Radford, Ilya Sutskever, and Dario Amodei.
\newblock Language models are few-shot learners.
\newblock In \emph{Advances in Neural Information Processing Systems 33: Annual Conference on Neural Information Processing Systems 2020, NeurIPS 2020, December 6-12, 2020, virtual}, 2020.

\bibitem[Cai et~al.(2023)Cai, Yin, Zeng, Wei, Sun, Wang, Pang, Mei, Zhang, Zhang, Loy, Yang, and Liu]{cai2023smplerx}
Zhongang Cai, Wanqi Yin, Ailing Zeng, Chen Wei, Qingping Sun, Yanjun Wang, Hui~En Pang, Haiyi Mei, Mingyuan Zhang, Lei Zhang, Chen~Change Loy, Lei Yang, and Ziwei Liu.
\newblock Smpler-x: Scaling up expressive human pose and shape estimation, 2023.

\bibitem[Camg{\"o}z et~al.(2018)Camg{\"o}z, Hadfield, Koller, Ney, and Bowden]{camgoz2018neural}
Necati~Cihan Camg{\"o}z, Simon Hadfield, Oscar Koller, Hermann Ney, and Richard Bowden.
\newblock {Neural Sign Language Translation}.
\newblock In \emph{Proceedings of the IEEE Conference on Computer Vision and Pattern Recognition (CVPR)}, 2018.

\bibitem[Cao et~al.(2017)Cao, Hidalgo, Simon, Wei, and Sheikh]{cao2018openpose}
Zhe Cao, Gines Hidalgo, Tomas Simon, Shih-En Wei, and Yaser Sheikh.
\newblock {OpenPose: Realtime Multi-Person 2D Pose Estimation using Part Affinity Fields}.
\newblock In \emph{Proceedings of the IEEE Conference on Computer Vision and Pattern Recognition (CVPR)}, 2017.

\bibitem[{Cao} et~al.(2019){Cao}, {Hidalgo Martinez}, {Simon}, {Wei}, and {Sheikh}]{openpose}
Z. {Cao}, G. {Hidalgo Martinez}, T. {Simon}, S. {Wei}, and Y.~A. {Sheikh}.
\newblock Openpose: Realtime multi-person 2d pose estimation using part affinity fields.
\newblock \emph{IEEE Transactions on Pattern Analysis and Machine Intelligence}, 2019.

\bibitem[Carreira and Zisserman(2017)]{carreira-2017-i3d}
J. Carreira and A. Zisserman.
\newblock Quo vadis, action recognition? a new model and the kinetics dataset.
\newblock In \emph{Proceedings of the IEEE/CVF Conference on Computer Vision and Pattern Recognition (CVPR)}, 2017.

\bibitem[Chan et~al.(2019)Chan, Ginosar, Zhou, and Efros]{chan2019everybody}
Caroline Chan, Shiry Ginosar, Tinghui Zhou, and Alexei~A Efros.
\newblock {Everybody Dance Now}.
\newblock In \emph{Proceedings of the IEEE International Conference on Computer Vision (CVPR)}, 2019.

\bibitem[Chen et~al.(2023)Chen, Jiang, Liu, Huang, Fu, Chen, and Yu]{chen2023executing}
Xin Chen, Biao Jiang, Wen Liu, Zilong Huang, Bin Fu, Tao Chen, and Gang Yu.
\newblock Executing your commands via motion diffusion in latent space.
\newblock In \emph{Proceedings of the IEEE/CVF Conference on Computer Vision and Pattern Recognition}, pages 18000--18010, 2023.

\bibitem[Chowdhery et~al.(2022)Chowdhery, Narang, Devlin, Bosma, Mishra, Roberts, Barham, Chung, Sutton, Gehrmann, Schuh, Shi, Tsvyashchenko, Maynez, Rao, Barnes, Tay, Shazeer, Prabhakaran, Reif, Du, Hutchinson, Pope, Bradbury, Austin, Isard, Gur{-}Ari, Yin, Duke, Levskaya, Ghemawat, Dev, Michalewski, Garcia, Misra, Robinson, Fedus, Zhou, Ippolito, Luan, Lim, Zoph, Spiridonov, Sepassi, Dohan, Agrawal, Omernick, Dai, Pillai, Pellat, Lewkowycz, Moreira, Child, Polozov, Lee, Zhou, Wang, Saeta, Diaz, Firat, Catasta, Wei, Meier{-}Hellstern, Eck, Dean, Petrov, and Fiedel]{Chowdhery-arxiv-2022-PaLM}
Aakanksha Chowdhery, Sharan Narang, Jacob Devlin, Maarten Bosma, Gaurav Mishra, Adam Roberts, Paul Barham, Hyung~Won Chung, Charles Sutton, Sebastian Gehrmann, Parker Schuh, Kensen Shi, Sasha Tsvyashchenko, Joshua Maynez, Abhishek Rao, Parker Barnes, Yi Tay, Noam Shazeer, Vinodkumar Prabhakaran, Emily Reif, Nan Du, Ben Hutchinson, Reiner Pope, James Bradbury, Jacob Austin, Michael Isard, Guy Gur{-}Ari, Pengcheng Yin, Toju Duke, Anselm Levskaya, Sanjay Ghemawat, Sunipa Dev, Henryk Michalewski, Xavier Garcia, Vedant Misra, Kevin Robinson, Liam Fedus, Denny Zhou, Daphne Ippolito, David Luan, Hyeontaek Lim, Barret Zoph, Alexander Spiridonov, Ryan Sepassi, David Dohan, Shivani Agrawal, Mark Omernick, Andrew~M. Dai, Thanumalayan~Sankaranarayana Pillai, Marie Pellat, Aitor Lewkowycz, Erica Moreira, Rewon Child, Oleksandr Polozov, Katherine Lee, Zongwei Zhou, Xuezhi Wang, Brennan Saeta, Mark Diaz, Orhan Firat, Michele Catasta, Jason Wei, Kathy Meier{-}Hellstern, Douglas Eck, Jeff Dean, Slav Petrov, and Noah Fiedel.
\newblock Palm: Scaling language modeling with pathways.
\newblock \emph{CoRR}, abs/2204.02311, 2022.

\bibitem[Cihan~Camgoz et~al.(2018)Cihan~Camgoz, Hadfield, Koller, Ney, and Bowden]{SLTranslation}
Necati Cihan~Camgoz, Simon Hadfield, Oscar Koller, Hermann Ney, and Richard Bowden.
\newblock Neural sign language translation.
\newblock In \emph{CVPR}, pages 7784--7793, 2018.

\bibitem[Cihan~Camg{\"o}z et~al.(2020)Cihan~Camg{\"o}z, Koller, Hadfield, and Bowden]{camgoz2020sign}
Necati Cihan~Camg{\"o}z, Oscar Koller, Simon Hadfield, and Richard Bowden.
\newblock {Sign Language Transformers: Joint End-to-end Sign Language Recognition and Translation}.
\newblock In \emph{Proceedings of the IEEE Conference on Computer Vision and Pattern Recognition (CVPR)}, 2020.

\bibitem[Cooper and Bowden(2007)]{cooper2007large}
Helen Cooper and Richard Bowden.
\newblock {Large Lexicon Detection of Sign Language}.
\newblock In \emph{International Workshop on Human-Computer Interaction}, 2007.

\bibitem[Cui et~al.(2017)Cui, Liu, and Zhang]{cui2017recurrent}
Runpeng Cui, Hu Liu, and Changshui Zhang.
\newblock {Recurrent Convolutional Neural Networks for Continuous Sign Language Recognition by Staged Optimization}.
\newblock In \emph{Proceedings of the IEEE Conference on Computer Vision and Pattern Recognition (CVPR)}, 2017.

\bibitem[Deng et~al.(2020)Deng, Yang, Chen, Wen, and Tong]{deng2020disentangled}
Yu Deng, Jiaolong Yang, Dong Chen, Fang Wen, and Xin Tong.
\newblock {Disentangled and Controllable Face Image Generation via 3D Imitative-Contrastive Learning}.
\newblock In \emph{Proceedings of the IEEE Conference on Computer Vision and Pattern Recognition (CVPR)}, 2020.

\bibitem[Duarte et~al.(2021{\natexlab{a}})Duarte, Palaskar, Ventura, Ghadiyaram, DeHaan, Metze, Torres, and Giro-i Nieto]{Duarte_2021_how2sign}
Amanda Duarte, Shruti Palaskar, Lucas Ventura, Deepti Ghadiyaram, Kenneth DeHaan, Florian Metze, Jordi Torres, and Xavier Giro-i Nieto.
\newblock {How2Sign: A Large-scale Multimodal Dataset for Continuous American Sign Language}.
\newblock In \emph{Proceedings of the IEEE/CVF Conference on Computer Vision and Pattern Recognition (CVPR)}, 2021{\natexlab{a}}.

\bibitem[Duarte et~al.(2021{\natexlab{b}})Duarte, Palaskar, Ventura, Ghadiyaram, DeHaan, Metze, Torres, and Giro-i Nieto]{duarte2021how2sign}
Amanda Duarte, Shruti Palaskar, Lucas Ventura, Deepti Ghadiyaram, Kenneth DeHaan, Florian Metze, Jordi Torres, and Xavier Giro-i Nieto.
\newblock {How2Sign: A Large-Scale Multimodal Dataset for Continuous American Sign Language}.
\newblock In \emph{Proceedings of the IEEE/CVF Conference on Computer Vision and Pattern Recognition (CVPR)}, 2021{\natexlab{b}}.

\bibitem[Ebling(2016)]{ebling2016automatic}
Sarah Ebling.
\newblock \emph{Automatic Translation from German to Synthesized Swiss German Sign Language}.
\newblock PhD thesis, University of Zurich, 2016.

\bibitem[Fang et~al.(2024)Fang, Sui, Zhou, Zhang, Zhong, Zhao, Tian, and Chen]{fang2024signdiffdiffusionmodelsamerican}
Sen Fang, Chunyu Sui, Yanghao Zhou, Xuedong Zhang, Hongbin Zhong, Minyu Zhao, Yapeng Tian, and Chen Chen.
\newblock Signdiff: Diffusion models for american sign language production, 2024.

\bibitem[Fang et~al.(2025)Fang, Sui, Yi, Neidle, and Metaxas]{fang2025signxfoundationmodelsign}
Sen Fang, Chunyu Sui, Hongwei Yi, Carol Neidle, and Dimitris~N. Metaxas.
\newblock Signx: The foundation model for sign recognition, 2025.

\bibitem[Feng et~al.(2023)Feng, Liu, Yu, Yao, Hui, Guo, Lin, Xue, Shi, Li, Li, Kang, Lei, Cui, Ren, and Xie]{feng2023dreamoving}
Mengyang Feng, Jinlin Liu, Kai Yu, Yuan Yao, Zheng Hui, Xiefan Guo, Xianhui Lin, Haolan Xue, Chen Shi, Xiaowen Li, Aojie Li, Xiaoyang Kang, Biwen Lei, Miaomiao Cui, Peiran Ren, and Xuansong Xie.
\newblock Dreamoving: A human video generation framework based on diffusion models, 2023.

\bibitem[Forster et~al.(2012)Forster, Schmidt, Hoyoux, Koller, Zelle, Piater, and Ney]{forster2012rwth}
Jens Forster, Christoph Schmidt, Thomas Hoyoux, Oscar Koller, Uwe Zelle, Justus~H Piater, and Hermann Ney.
\newblock {RWTH-PHOENIX-Weather: A Large Vocabulary Sign Language Recognition and Translation Corpus}.
\newblock In \emph{Proceedings of the International Conference on Language Resources and Evaluation (LREC)}, 2012.

\bibitem[Forster et~al.(2014)Forster, Schmidt, Koller, Bellgardt, and Ney]{RWTH-PHOENIX-Weather-2014}
Jens Forster, Christoph Schmidt, Oscar Koller, Martin Bellgardt, and Hermann Ney.
\newblock Extensions of the sign language recognition and translation corpus rwth-phoenix-weather.
\newblock In \emph{Proceedings of the Ninth International Conference on Language Resources and Evaluation ({LREC}'14)}, pages 1911--1916, 2014.

\bibitem[Gong et~al.(2024)Gong, Foo, He, Rahmani, and Liu]{gong2024llms}
Jia Gong, Lin~Geng Foo, Yixuan He, Hossein Rahmani, and Jun Liu.
\newblock Llms are good sign language translators, 2024.

\bibitem[Goodfellow et~al.(2014)Goodfellow, Pouget-Abadie, Mirza, Xu, Warde-Farley, Ozair, Courville, and Bengio]{goodfellow2014generative}
Ian Goodfellow, Jean Pouget-Abadie, Mehdi Mirza, Bing Xu, David Warde-Farley, Sherjil Ozair, Aaron Courville, and Yoshua Bengio.
\newblock {Generative Adversarial Nets}.
\newblock In \emph{Proceedings of the Advances in Neural Information Processing Systems (NIPS)}, 2014.

\bibitem[Grobel and Assan(1997)]{grobel1997isolated}
Kirsti Grobel and Marcell Assan.
\newblock {Isolated Sign Language Recognition using Hidden Markov Models}.
\newblock In \emph{IEEE International Conference on Systems, Man, and Cybernetics}, 1997.

\bibitem[Gueuwou et~al.(2023)Gueuwou, Siake, Leong, and M{\"u}ller]{gueuwou2023jwsign}
Shester Gueuwou, Sophie Siake, Colin Leong, and Mathias M{\"u}ller.
\newblock Jwsign: A highly multilingual corpus of bible translations for more diversity in sign language processing.
\newblock \emph{arXiv preprint arXiv:2311.10174}, 2023.

\bibitem[G{\"u}ler et~al.(2018)G{\"u}ler, Neverova, and Kokkinos]{guler2018densepose}
R{\i}za~Alp G{\"u}ler, Natalia Neverova, and Iasonas Kokkinos.
\newblock Densepose: Dense human pose estimation in the wild.
\newblock In \emph{Proceedings of the IEEE Conference on Computer Vision and Pattern Recognition}, pages 7297--7306, 2018.

\bibitem[Ham et~al.(2021)Ham, Park, Jang, Oh, Yun, Yoon, Kim, Park, and Kweon]{Ham21}
Soomin Ham, Kibaek Park, YeongJun Jang, Youngtaek Oh, Seokmin Yun, Sukwon Yoon, Chango~Jo Kim, Han-Mu Park, and In~So Kweon.
\newblock Ksl-guide: A large-scale korean sign language dataset including interrogative sentences for guiding the deaf and hard-of-hearing.
\newblock In \emph{IEEE International Conference on Automatic Face and Gesture Recognition}, 2021.

\bibitem[Hanke et~al.(2020)Hanke, Schulder, Konrad, and Jahn]{DGS-Korpus}
Thomas Hanke, Marc Schulder, Reiner Konrad, and Elena Jahn.
\newblock Extending the public dgs corpus in size and depth.
\newblock In \emph{LREC2020 - Workshop on the Representation and Processing of Sign Languages}, pages 75--82, 2020.

\bibitem[Hilzensauer and Krammer(2015)]{hilzensauer2015multilingual}
Marlene Hilzensauer and Klaudia Krammer.
\newblock A multilingual dictionary for sign languages:" spreadthesign".
\newblock In \emph{ICERI2015 Proceedings}, pages 7826--7834. IATED, 2015.

\bibitem[Hoffmann et~al.(2022)Hoffmann, Borgeaud, Mensch, Buchatskaya, Cai, Rutherford, de~Las~Casas, Hendricks, Welbl, Clark, Hennigan, Noland, Millican, van~den Driessche, Damoc, Guy, Osindero, Simonyan, Elsen, Rae, Vinyals, and Sifre]{Hoffmann-arxiv-2022-Training}
Jordan Hoffmann, Sebastian Borgeaud, Arthur Mensch, Elena Buchatskaya, Trevor Cai, Eliza Rutherford, Diego de Las~Casas, Lisa~Anne Hendricks, Johannes Welbl, Aidan Clark, Tom Hennigan, Eric Noland, Katie Millican, George van~den Driessche, Bogdan Damoc, Aurelia Guy, Simon Osindero, Karen Simonyan, Erich Elsen, Jack~W. Rae, Oriol Vinyals, and Laurent Sifre.
\newblock Training compute-optimal large language models.
\newblock abs/2203.15556, 2022.

\bibitem[Hu et~al.(2021)Hu, Shen, Wallis, Allen-Zhu, Li, Wang, Wang, and Chen]{hu2021lora}
Edward~J Hu, Yelong Shen, Phillip Wallis, Zeyuan Allen-Zhu, Yuanzhi Li, Shean Wang, Lu Wang, and Weizhu Chen.
\newblock Lora: Low-rank adaptation of large language models.
\newblock \emph{arXiv preprint arXiv:2106.09685}, 2021.

\bibitem[Huang et~al.(2018)Huang, Zhou, Zhang, Li, and Li]{video-based}
Jie Huang, Wengang Zhou, Qilin Zhang, Houqiang Li, and Weiping Li.
\newblock Video-based sign language recognition without temporal segmentation.
\newblock In \emph{AAAI}, 2018.

\bibitem[Huang et~al.(2021)Huang, Pan, Zhao, and Tian]{huang2021towards}
Wencan Huang, Wenwen Pan, Zhou Zhao, and Qi Tian.
\newblock {Towards Fast and High-Quality Sign Language Production}.
\newblock In \emph{Proceedings of the 29th ACM International Conference on Multimedia}, 2021.

\bibitem[Hwang et~al.(2021)Hwang, Kim, and Park]{hwang2021non}
Eui~Jun Hwang, Jung-Ho Kim, and Jong~C. Park.
\newblock Non-autoregressive sign language production with gaussian space.
\newblock In \emph{The 32nd British Machine Vision Conference (BMVC 21)}. British Machine Vision Conference (BMVC), 2021.

\bibitem[Isola et~al.(2017)Isola, Zhu, Zhou, and Efros]{isola2017image}
Phillip Isola, Jun-Yan Zhu, Tinghui Zhou, and Alexei~A Efros.
\newblock {Image-to-Image Translation with Conditional Adversarial Networks}.
\newblock In \emph{Proceedings of the IEEE Conference on Computer Vision and Pattern Recognition (CVPR)}, 2017.

\bibitem[Kadir et~al.(2004)Kadir, Bowden, Ong, and Zisserman]{kadir2004minimal}
Timor Kadir, Richard Bowden, Eng-Jon Ong, and Andrew Zisserman.
\newblock {Minimal Training, Large Lexicon, Unconstrained Sign Language Recognition}.
\newblock In \emph{Proceedings of the British Machine Vision Conference (BMVC)}, 2004.

\bibitem[Kaplan et~al.(2020)Kaplan, McCandlish, Henighan, Brown, Chess, Child, Gray, Radford, Wu, and Amodei]{Kaplan-arxiv-2020-Scaling}
Jared Kaplan, Sam McCandlish, Tom Henighan, Tom~B. Brown, Benjamin Chess, Rewon Child, Scott Gray, Alec Radford, Jeffrey Wu, and Dario Amodei.
\newblock Scaling laws for neural language models.
\newblock \emph{CoRR}, abs/2001.08361, 2020.

\bibitem[Kayahan and Güngör(2019)]{8778347}
Dilek Kayahan and Tunga Güngör.
\newblock A hybrid translation system from turkish spoken language to turkish sign language.
\newblock In \emph{2019 IEEE International Symposium on INnovations in Intelligent SysTems and Applications (INISTA)}, pages 1--6, 2019.

\bibitem[Ko et~al.(2019)Ko, Kim, Jung, and Cho]{ko2019neural}
Sang-Ki Ko, Chang~Jo Kim, Hyedong Jung, and Choongsang Cho.
\newblock {Neural Sign Language Translation based on Human Keypoint Estimation}.
\newblock \emph{Applied Sciences}, 2019.

\bibitem[Koller(2020)]{koller2020quantitative}
Oscar Koller.
\newblock {Quantitative Survey of the State of the Art in Sign Language Recognition}.
\newblock \emph{arXiv preprint arXiv:2008.09918}, 2020.

\bibitem[Koller et~al.(2015{\natexlab{a}})Koller, Forster, and Ney]{koller15:cslr}
Oscar Koller, Jens Forster, and Hermann Ney.
\newblock Continuous sign language recognition: Towards large vocabulary statistical recognition systems handling multiple signers.
\newblock \emph{Computer Vision and Image Understanding}, 141:\penalty0 108--125, 2015{\natexlab{a}}.

\bibitem[Koller et~al.(2015{\natexlab{b}})Koller, Forster, and Ney]{koller2015continuous}
Oscar Koller, Jens Forster, and Hermann Ney.
\newblock {Continuous Sign Language Recognition: Towards Large Vocabulary Statistical Recognition Systems Handling Multiple Signers}.
\newblock \emph{Computer Vision and Image Understanding (CVIU)}, 2015{\natexlab{b}}.

\bibitem[Kowalski et~al.(2020)Kowalski, Garbin, Estellers, Baltru{\v{s}}aitis, Johnson, and Shotton]{kowalski2020config}
Marek Kowalski, Stephan~J Garbin, Virginia Estellers, Tadas Baltru{\v{s}}aitis, Matthew Johnson, and Jamie Shotton.
\newblock {CONFIG: Controllable Neural Face Image Generation}.
\newblock In \emph{Proceedings of the European Conference on Computer Vision (ECCV)}, 2020.

\bibitem[Kreutzer et~al.(2019)Kreutzer, Bastings, and Riezler]{JoeyNMT}
Julia Kreutzer, Joost Bastings, and Stefan Riezler.
\newblock {Joey {NMT}: A Minimalist {NMT} Toolkit for Novices}.
\newblock In \emph{Proceedings of the Conference on Empirical Methods in Natural Language Processing (EMNLP)}, 2019.

\bibitem[Lin(2004)]{lin-acl-2004-rouge}
Chin-Yew Lin.
\newblock {ROUGE}: A package for automatic evaluation of summaries.
\newblock In \emph{Text Summarization Branches Out}, pages 74--81. Association for Computational Linguistics, 2004.

\bibitem[Liu et~al.(2019)Liu, De~Nadai, Zen, Sebe, and Lepri]{liu2019gesture}
Yahui Liu, Marco De~Nadai, Gloria Zen, Nicu Sebe, and Bruno Lepri.
\newblock {Gesture-to-Gesture Translation in the Wild via Category-Independent Conditional Maps}.
\newblock In \emph{Proceedings of the 27th ACM International Conference on Multimedia}, 2019.

\bibitem[Loper et~al.(2015)Loper, Mahmood, Romero, Pons-Moll, and Black]{smpl}
Matthew Loper, Naureen Mahmood, Javier Romero, Gerard Pons-Moll, and Michael~J. Black.
\newblock {SMPL}: {A} skinned multi-person linear model.
\newblock In \emph{ACM TOG}, 2015.

\bibitem[Lugaresi et~al.(2019)Lugaresi, Tang, Nash, McClanahan, Uboweja, Hays, Zhang, Chang, Yong, Lee, et~al.]{MediaPipe}
Camillo Lugaresi, Jiuqiang Tang, Hadon Nash, Chris McClanahan, Esha Uboweja, Michael Hays, Fan Zhang, Chuo-Ling Chang, Ming~Guang Yong, Juhyun Lee, et~al.
\newblock Mediapipe: A framework for building perception pipelines.
\newblock \emph{arXiv preprint arXiv:1906.08172}, 2019.

\bibitem[Ma et~al.(2017)Ma, Jia, Sun, Schiele, Tuytelaars, and Van~Gool]{ma2017pose}
Liqian Ma, Xu Jia, Qianru Sun, Bernt Schiele, Tinne Tuytelaars, and Luc Van~Gool.
\newblock {Pose Guided Person Image Generation}.
\newblock In \emph{Advances in Neural Information Processing Systems (NIPS)}, 2017.

\bibitem[Mallya et~al.(2020)Mallya, Wang, Sapra, and Liu]{mallya2020world}
Arun Mallya, Ting-Chun Wang, Karan Sapra, and Ming-Yu Liu.
\newblock {World-Consistent Video-to-Video Synthesis}.
\newblock In \emph{Proceedings of the European Conference on Computer Vision (ECCV)}, 2020.

\bibitem[Matthes et~al.(2012)Matthes, Hanke, Regen, Storz, Worseck, Efthimiou, Dimou, Braffort, Glauert, and Safar]{matthes2012dicta}
Silke Matthes, Thomas Hanke, Anja Regen, Jakob Storz, Satu Worseck, Eleni Efthimiou, Athanasia-Lida Dimou, Annelies Braffort, John Glauert, and Eva Safar.
\newblock Dicta-sign-building a multilingual sign language corpus.
\newblock In \emph{5th Workshop on the Representation and Processing of Sign languages: Interactions between Corpus and Lexicon. Satellite Workshop to the eighth International Conference on language Resources and Evaluation (LREC-2012)}, 2012.

\bibitem[Men et~al.(2020)Men, Mao, Jiang, Ma, and Lian]{men2020controllable}
Yifang Men, Yiming Mao, Yuning Jiang, Wei-Ying Ma, and Zhouhui Lian.
\newblock {Controllable Person Image Synthesis with Attribute-Decomposed GAN}.
\newblock In \emph{Proceedings of the IEEE Conference on Computer Vision and Pattern Recognition (CVPR)}, 2020.

\bibitem[Moryossef and M\"{u}ller(2021)]{moryossef2021datasets}
Amit Moryossef and Mathias M\"{u}ller.
\newblock Sign language datasets.
\newblock \url{https://github.com/sign-language-processing/datasets}, 2021.

\bibitem[M{\"u}ller et~al.(2023)M{\"u}ller, Alikhani, Avramidis, Bowden, Braffort, Cihan~Camg{\"o}z, Ebling, Espa{\~n}a-Bonet, G{\"o}hring, Grundkiewicz, Inan, Jiang, Koller, Moryossef, Rios, Shterionov, Sidler-Miserez, Tissi, and Van~Landuyt]{muller-etal-2023-findings}
Mathias M{\"u}ller, Malihe Alikhani, Eleftherios Avramidis, Richard Bowden, Annelies Braffort, Necati Cihan~Camg{\"o}z, Sarah Ebling, Cristina Espa{\~n}a-Bonet, Anne G{\"o}hring, Roman Grundkiewicz, Mert Inan, Zifan Jiang, Oscar Koller, Amit Moryossef, Annette Rios, Dimitar Shterionov, Sandra Sidler-Miserez, Katja Tissi, and Davy Van~Landuyt.
\newblock Findings of the second {WMT} shared task on sign language translation ({WMT}-{SLT}23).
\newblock In \emph{Proceedings of the Eighth Conference on Machine Translation}, pages 68--94, Singapore, 2023. Association for Computational Linguistics.

\bibitem[Neidle and Vogler(2012)]{NCSLGR}
Carol Neidle and Christian Vogler.
\newblock A new web interface to facilitate access to corpora: Development of the asllrp data access interface (dai).
\newblock In \emph{Proc. 5th Workshop on the Representation and Processing of Sign Languages: Interactions between Corpus and Lexicon, LREC}, 2012.

\bibitem[Papineni et~al.(2002)Papineni, Roukos, Ward, and Zhu]{bleu}
Kishore Papineni, Salim Roukos, Todd Ward, and Wei-Jing Zhu.
\newblock {B}leu: a method for automatic evaluation of machine translation.
\newblock In \emph{Proceedings of the 40th Annual Meeting of the Association for Computational Linguistics}, pages 311--318, Philadelphia, Pennsylvania, USA, 2002. Association for Computational Linguistics.

\bibitem[Paszke et~al.(2017)Paszke, Gross, Chintala, Chanan, Yang, DeVito, Lin, Desmaison, Antiga, and Lerer]{paszke2017automatic}
Adam Paszke, Sam Gross, Soumith Chintala, Gregory Chanan, Edward Yang, Zachary DeVito, Zeming Lin, Alban Desmaison, Luca Antiga, and Adam Lerer.
\newblock {Automatic Differentiation in PyTorch}.
\newblock In \emph{NIPS Autodiff Workshop}, 2017.

\bibitem[Prajwal et~al.(2020)Prajwal, Mukhopadhyay, Namboodiri, and Jawahar]{10.1145/3394171.3413532}
K~R Prajwal, Rudrabha Mukhopadhyay, Vinay~P. Namboodiri, and C.V. Jawahar.
\newblock A lip sync expert is all you need for speech to lip generation in the wild.
\newblock In \emph{Proceedings of the 28th ACM International Conference on Multimedia}, page 484–492, New York, NY, USA, 2020. Association for Computing Machinery.

\bibitem[Qiu et~al.(2021)Qiu, Anwar, and Barnes]{qiu2021dense}
Shi Qiu, Saeed Anwar, and Nick Barnes.
\newblock Dense-resolution network for point cloud classification and segmentation.
\newblock In \emph{WACV}, pages 3813--3822, 2021.

\bibitem[Radford et~al.(2015)Radford, Metz, and Chintala]{radford2015unsupervised}
Alec Radford, Luke Metz, and Soumith Chintala.
\newblock {Unsupervised Representation Learning with Deep Convolutional Generative Adversarial Networks}.
\newblock \emph{arXiv preprint arXiv:1511.06434}, 2015.

\bibitem[Radford et~al.(2019)Radford, Wu, Child, Luan, Amodei, Sutskever, et~al.]{radford-blog-2019-language}
Alec Radford, Jeffrey Wu, Rewon Child, David Luan, Dario Amodei, Ilya Sutskever, et~al.
\newblock Language models are unsupervised multitask learners.
\newblock \emph{OpenAI blog}, page~9, 2019.

\bibitem[Raffel et~al.(2020)Raffel, Shazeer, Roberts, Lee, Narang, Matena, Zhou, Li, and Liu]{raffel2020exploring}
Colin Raffel, Noam Shazeer, Adam Roberts, Katherine Lee, Sharan Narang, Michael Matena, Yanqi Zhou, Wei Li, and Peter~J Liu.
\newblock Exploring the limits of transfer learning with a unified text-to-text transformer.
\newblock \emph{The Journal of Machine Learning Research}, 21\penalty0 (1):\penalty0 5485--5551, 2020.

\bibitem[Rombach et~al.(2022)Rombach, Blattmann, Lorenz, Esser, and Ommer]{Rombach_2022_CVPR}
Robin Rombach, Andreas Blattmann, Dominik Lorenz, Patrick Esser, and Bj\"orn Ommer.
\newblock High-resolution image synthesis with latent diffusion models.
\newblock In \emph{Proceedings of the IEEE/CVF Conference on Computer Vision and Pattern Recognition (CVPR)}, pages 10684--10695, 2022.

\bibitem[Ronchetti et~al.(2016)Ronchetti, Quiroga, Estrebou, Lanzarini, and Rosete]{Ronchetti2016}
Franco Ronchetti, Facundo Quiroga, Cesar Estrebou, Laura Lanzarini, and Alejandro Rosete.
\newblock Lsa64: A dataset of argentinian sign language.
\newblock \emph{XX II Congreso Argentino de Ciencias de la Computación (CACIC)}, 2016.

\bibitem[Saharia et~al.(2022)Saharia, Chan, Saxena, Li, Whang, Denton, Ghasemipour, Gontijo~Lopes, Karagol~Ayan, Salimans, Ho, Fleet, and Norouzi]{NEURIPS2022_ec795aea}
Chitwan Saharia, William Chan, Saurabh Saxena, Lala Li, Jay Whang, Emily~L Denton, Kamyar Ghasemipour, Raphael Gontijo~Lopes, Burcu Karagol~Ayan, Tim Salimans, Jonathan Ho, David~J Fleet, and Mohammad Norouzi.
\newblock Photorealistic text-to-image diffusion models with deep language understanding.
\newblock In \emph{Advances in Neural Information Processing Systems}, pages 36479--36494. Curran Associates, Inc., 2022.

\bibitem[Saunders et~al.(2020{\natexlab{a}})Saunders, Camg{\"o}z, and Bowden]{saunders2020adversarial}
Ben Saunders, Necati~Cihan Camg{\"o}z, and Richard Bowden.
\newblock {Adversarial Training for Multi-Channel Sign Language Production}.
\newblock In \emph{Proceedings of the British Machine Vision Conference (BMVC)}, 2020{\natexlab{a}}.

\bibitem[Saunders et~al.(2020{\natexlab{b}})Saunders, Camg{\"o}z, and Bowden]{saunders2020progressive}
Ben Saunders, Necati~Cihan Camg{\"o}z, and Richard Bowden.
\newblock {Progressive Transformers for End-to-End Sign Language Production}.
\newblock In \emph{Proceedings of the European Conference on Computer Vision (ECCV)}, 2020{\natexlab{b}}.

\bibitem[Saunders et~al.(2021{\natexlab{a}})Saunders, Camg{\"o}z, and Bowden]{saunders2021continuous}
Ben Saunders, Necati~Cihan Camg{\"o}z, and Richard Bowden.
\newblock {Continuous 3D Multi-Channel Sign Language Production via Progressive Transformers and Mixture Density Networks}.
\newblock \emph{International Journal of Computer Vision (IJCV)}, 2021{\natexlab{a}}.

\bibitem[Saunders et~al.(2021{\natexlab{b}})Saunders, Camg{\"o}z, and Bowden]{saunders2021mixed}
Ben Saunders, Necati~Cihan Camg{\"o}z, and Richard Bowden.
\newblock {Mixed SIGNals: Sign Language Production via a Mixture of Motion Primitives}.
\newblock In \emph{Proceedings of the International Conference on Computer Vision (ICCV)}, 2021{\natexlab{b}}.

\bibitem[Saunders et~al.(2021{\natexlab{c}})Saunders, Camgoz, and Bowden]{saunders2021skeletal}
Ben Saunders, Necati~Cihan Camgoz, and Richard Bowden.
\newblock {Skeletal Graph Self-Attention: Embedding a Skeleton Inductive Bias into Sign Language Production}.
\newblock \emph{arXiv preprint arXiv:2112.05277}, 2021{\natexlab{c}}.

\bibitem[Saunders et~al.(2022)Saunders, Camgoz, and Bowden]{Saunders_2022_CVPR}
Ben Saunders, Necati~Cihan Camgoz, and Richard Bowden.
\newblock Signing at scale: Learning to co-articulate signs for large-scale photo-realistic sign language production.
\newblock In \emph{Proceedings of the IEEE/CVF Conference on Computer Vision and Pattern Recognition (CVPR)}, pages 5141--5151, 2022.

\bibitem[Schembri et~al.(2013)Schembri, Fenlon, Rentelis, Reynolds, and Cormier]{bsl-corpus}
Adam Schembri, Jordan Fenlon, Ramas Rentelis, Sally Reynolds, and Kearsy Cormier.
\newblock Building the british sign language corpus.
\newblock \emph{Language Documentation \& Conservation}, 7:\penalty0 136--154, 2013.

\bibitem[Shanahan(2022)]{Shanahan-arxiv-2022-Talking}
Murray Shanahan.
\newblock Talking about large language models.
\newblock \emph{CoRR}, abs/2212.03551, 2022.

\bibitem[Shi et~al.(2022)Shi, Brentari, Shakhnarovich, and Livescu]{shi-etal-2022-openASL}
Bowen Shi, Diane Brentari, Greg Shakhnarovich, and Karen Livescu.
\newblock Open-domain sign language translation learned from online video.
\newblock \emph{Proceedings of the 2022 Conference on Empirical Methods in Natural Language Processing}, 2022.

\bibitem[Siarohin et~al.(2018)Siarohin, Sangineto, Lathuiliere, and Sebe]{siarohin2018deformable}
Aliaksandr Siarohin, Enver Sangineto, St{\'e}phane Lathuiliere, and Nicu Sebe.
\newblock {Deformable GANs for Pose-Based Human Image Generation}.
\newblock In \emph{Proceedings of the IEEE Conference on Computer Vision and Pattern Recognition (CVPR)}, 2018.

\bibitem[Sincan and Keles(2020)]{9210578}
Ozge~Mercanoglu Sincan and Hacer~Yalim Keles.
\newblock Autsl: A large scale multi-modal turkish sign language dataset and baseline methods.
\newblock \emph{IEEE Access}, 8:\penalty0 181340--181355, 2020.

\bibitem[Stoll et~al.(2018)Stoll, Camg{\"o}z, Hadfield, and Bowden]{stoll2018sign}
Stephanie Stoll, Necati~Cihan Camg{\"o}z, Simon Hadfield, and Richard Bowden.
\newblock {Sign Language Production using Neural Machine Translation and Generative Adversarial Networks}.
\newblock In \emph{Proceedings of the British Machine Vision Conference (BMVC)}, 2018.

\bibitem[Stoll et~al.(2020)Stoll, Camg{\"o}z, Hadfield, and Bowden]{stoll2020text2sign}
Stephanie Stoll, Necati~Cihan Camg{\"o}z, Simon Hadfield, and Richard Bowden.
\newblock {Text2Sign: Towards Sign Language Production using Neural Machine Translation and Generative Adversarial Networks}.
\newblock \emph{International Journal of Computer Vision (IJCV)}, 2020.

\bibitem[Szegedy et~al.(2015)Szegedy, Liu, Jia, Sermanet, Reed, Anguelov, Erhan, Vanhoucke, and Rabinovich]{7298594}
Christian Szegedy, Wei Liu, Yangqing Jia, Pierre Sermanet, Scott Reed, Dragomir Anguelov, Dumitru Erhan, Vincent Vanhoucke, and Andrew Rabinovich.
\newblock Going deeper with convolutions.
\newblock In \emph{2015 IEEE Conference on Computer Vision and Pattern Recognition (CVPR)}, pages 1--9, 2015.

\bibitem[Tang et~al.(2018)Tang, Wang, Xu, Yan, and Sebe]{tang2018gesturegan}
Hao Tang, Wei Wang, Dan Xu, Yan Yan, and Nicu Sebe.
\newblock {GestureGAN for Hand Gesture-to-Gesture Translation in the wild}.
\newblock In \emph{Proceedings of the 26th ACM International Conference on Multimedia}, 2018.

\bibitem[Tang et~al.(2020)Tang, Bai, Zhang, Torr, and Sebe]{tang2020xinggan}
Hao Tang, Song Bai, Li Zhang, Philip~HS Torr, and Nicu Sebe.
\newblock {XingGAN for Person Image Generation}.
\newblock In \emph{Proceedings of the European Conference on Computer Vision (ECCV)}, 2020.

\bibitem[Tarrés et~al.(2023)Tarrés, Gállego, Duarte, Torres, and i~Nieto]{slt-how2sign-wicv2023}
Laia Tarrés, Gerard~I. Gállego, Amanda Duarte, Jordi Torres, and Xavier~Giró i Nieto.
\newblock Sign language translation from instructional videos.
\newblock In \emph{Proceedings of the IEEE/CVF Conference on Computer Vision and Pattern Recognition (CVPR) Workshops}, 2023.

\bibitem[Taylor et~al.(2022)Taylor, Kardas, Cucurull, Scialom, Hartshorn, Saravia, Poulton, Kerkez, and Stojnic]{Taylor-arxiv-2022-Galactica}
Ross Taylor, Marcin Kardas, Guillem Cucurull, Thomas Scialom, Anthony Hartshorn, Elvis Saravia, Andrew Poulton, Viktor Kerkez, and Robert Stojnic.
\newblock Galactica: {A} large language model for science.
\newblock \emph{CoRR}, abs/2211.09085, 2022.

\bibitem[Thies et~al.(2020)Thies, Elgharib, Tewari, Theobalt, and Nie{\ss}ner]{10.1007/978-3-030-58517-4_42}
Justus Thies, Mohamed Elgharib, Ayush Tewari, Christian Theobalt, and Matthias Nie{\ss}ner.
\newblock Neural voice puppetry: Audio-driven facial reenactment.
\newblock In \emph{Computer Vision -- ECCV 2020}, pages 716--731, Cham, 2020. Springer International Publishing.

\bibitem[Touvron et~al.(2023)Touvron, Lavril, Izacard, Martinet, Lachaux, Lacroix, Rozi{\`{e}}re, Goyal, Hambro, Azhar, Rodriguez, Joulin, Grave, and Lample]{Touvron-arxiv-2023-LLaMA}
Hugo Touvron, Thibaut Lavril, Gautier Izacard, Xavier Martinet, Marie{-}Anne Lachaux, Timoth{\'{e}}e Lacroix, Baptiste Rozi{\`{e}}re, Naman Goyal, Eric Hambro, Faisal Azhar, Aur{\'{e}}lien Rodriguez, Armand Joulin, Edouard Grave, and Guillaume Lample.
\newblock Llama: Open and efficient foundation language models.
\newblock \emph{CoRR}, 2023.

\bibitem[Tulyakov et~al.(2018)Tulyakov, Liu, Yang, and Kautz]{tulyakov2018mocogan}
Sergey Tulyakov, Ming-Yu Liu, Xiaodong Yang, and Jan Kautz.
\newblock {MoCoGAN: Decomposing Motion and Content for Video Generation}.
\newblock In \emph{Proceedings of the IEEE Conference on Computer Vision and Pattern Recognition (CVPR)}, 2018.

\bibitem[Ventura et~al.(2020)Ventura, Duarte, and Gir{\'o}-i Nieto]{ventura2020can}
Lucas Ventura, Amanda Duarte, and Xavier Gir{\'o}-i Nieto.
\newblock {Can Everybody Sign Now? Exploring Sign Language Video Generation from 2D Poses}.
\newblock In \emph{ECCV Sign Language Recognition, Translation and Production Workshop}, 2020.

\bibitem[Von~Agris and Kraiss(2010)]{SIGNUM}
U. Von~Agris and K.-F. Kraiss.
\newblock Signum database: Video corpus for signer-independent continuous sign language recognition.
\newblock In \emph{Workshop on Representation and Processing of Sign Languages}, pages 243--246, 2010.

\bibitem[Vondrick et~al.(2016)Vondrick, Pirsiavash, and Torralba]{vondrick2016generating}
Carl Vondrick, Hamed Pirsiavash, and Antonio Torralba.
\newblock {Generating Videos with Scene Dynamics}.
\newblock In \emph{Advances in Neural Information Processing Systems (NIPS)}, 2016.

\bibitem[Wang et~al.(2018{\natexlab{a}})Wang, Liu, Zhu, Liu, Tao, Kautz, and Catanzaro]{wang2018video}
Ting-Chun Wang, Ming-Yu Liu, Jun-Yan Zhu, Guilin Liu, Andrew Tao, Jan Kautz, and Bryan Catanzaro.
\newblock {Video-to-Video Synthesis}.
\newblock In \emph{Advances in Neural Information Processing Systems (NIPS)}, 2018{\natexlab{a}}.

\bibitem[Wang et~al.(2018{\natexlab{b}})Wang, Liu, Zhu, Tao, Kautz, and Catanzaro]{wang2018high}
Ting-Chun Wang, Ming-Yu Liu, Jun-Yan Zhu, Andrew Tao, Jan Kautz, and Bryan Catanzaro.
\newblock {High-Resolution Image Synthesis and Semantic Manipulation with Conditional GANs}.
\newblock In \emph{Proceedings of the IEEE Conference on Computer Vision and Pattern Recognition (CVPR)}, 2018{\natexlab{b}}.

\bibitem[Wang et~al.(2019)Wang, Liu, Tao, Liu, Kautz, and Catanzaro]{wang2019few}
Ting-Chun Wang, Ming-Yu Liu, Andrew Tao, Guilin Liu, Jan Kautz, and Bryan Catanzaro.
\newblock {Few-shot Video-to-Video Synthesis}.
\newblock In \emph{Advances in Neural Information Processing Systems (NeurIPS)}, 2019.

\bibitem[Wei et~al.(2020)Wei, Xu, Shen, and Huang]{wei2020gac}
Dongxu Wei, Xiaowei Xu, Haibin Shen, and Kejie Huang.
\newblock {GAC-GAN: A General Method for Appearance-Controllable Human Video Motion Transfer}.
\newblock \emph{IEEE Transactions on Multimedia}, 2020.

\bibitem[Wu et~al.(2020)Wu, Hoang, Lin, Xie, Chen, Lin, Wang, and Fan]{wu2020mm}
Zhenyu Wu, Duc Hoang, Shih-Yao Lin, Yusheng Xie, Liangjian Chen, Yen-Yu Lin, Zhangyang Wang, and Wei Fan.
\newblock {MM-Hand: 3D-Aware Multi-Modal Guided Hand Generative Network for 3D Hand Pose Synthesis}.
\newblock \emph{arXiv preprint arXiv:2010.01158}, 2020.

\bibitem[Xue et~al.(2020)Xue, Constant, Roberts, Kale, Al-Rfou, Siddhant, Barua, and Raffel]{xue2020mt5}
Linting Xue, Noah Constant, Adam Roberts, Mihir Kale, Rami Al-Rfou, Aditya Siddhant, Aditya Barua, and Colin Raffel.
\newblock mt5: A massively multilingual pre-trained text-to-text transformer.
\newblock \emph{arXiv preprint arXiv:2010.11934}, 2020.

\bibitem[Yang et~al.(2019)Yang, Jung, Kang, and Kim]{yang2019korean}
Seunghan Yang, Seungjun Jung, Heekwang Kang, and Changick Kim.
\newblock The korean sign language dataset for action recognition.
\newblock In \emph{International conference on multimedia modeling}, pages 532--542. Springer, 2019.

\bibitem[Yin et~al.(2022)Yin, Zhao, Jin, Zhang, Zeng, and He]{yin2022mlslt}
Aoxiong Yin, Zhou Zhao, Weike Jin, Meng Zhang, Xingshan Zeng, and Xiaofei He.
\newblock Mlslt: Towards multilingual sign language translation.
\newblock In \emph{Proceedings of the IEEE/CVF Conference on Computer Vision and Pattern Recognition}, pages 5109--5119, 2022.

\bibitem[Yu et~al.(2021)Yu, Yu, Li, and Ling]{8995571}
Lingyun Yu, Jun Yu, Mengyan Li, and Qiang Ling.
\newblock Multimodal inputs driven talking face generation with spatial–temporal dependency.
\newblock \emph{IEEE Transactions on Circuits and Systems for Video Technology}, 31\penalty0 (1):\penalty0 203--216, 2021.

\bibitem[Zakharov et~al.(2019)Zakharov, Shysheya, Burkov, and Lempitsky]{zakharov2019few}
Egor Zakharov, Aliaksandra Shysheya, Egor Burkov, and Victor Lempitsky.
\newblock {Few-Shot Adversarial Learning of Realistic Neural Talking Head Models}.
\newblock In \emph{Proceedings of the IEEE International Conference on Computer Vision (CVPR)}, 2019.

\bibitem[Zelinka and Kanis(2020)]{zelinka2020neural}
Jan Zelinka and Jakub Kanis.
\newblock {Neural Sign Language Synthesis: Words Are Our Glosses}.
\newblock In \emph{The IEEE Winter Conference on Applications of Computer Vision (WACV)}, 2020.

\bibitem[Zhang and Agrawala(2023)]{zhang2023adding}
Lvmin Zhang and Maneesh Agrawala.
\newblock Adding conditional control to text-to-image diffusion models, 2023.

\bibitem[Zhang et~al.(2022)Zhang, Yuan, Liao, and Zhang]{9747380}
Sibo Zhang, Jiahong Yuan, Miao Liao, and Liangjun Zhang.
\newblock Text2video: Text-driven talking-head video synthesis with personalized phoneme - pose dictionary.
\newblock In \emph{ICASSP 2022 - 2022 IEEE International Conference on Acoustics, Speech and Signal Processing (ICASSP)}, pages 2659--2663, 2022.

\bibitem[Zhou et~al.(2019)Zhou, Wang, Fang, Bui, and Berg]{zhou2019dance}
Yipin Zhou, Zhaowen Wang, Chen Fang, Trung Bui, and Tamara Berg.
\newblock {Dance Dance Generation: Motion Transfer for Internet Videos}.
\newblock In \emph{Proceedings of the IEEE International Conference on Computer Vision Workshops}, 2019.

\bibitem[Zhu et~al.(2017)Zhu, Park, Isola, and Efros]{zhu2017unpaired}
Jun-Yan Zhu, Taesung Park, Phillip Isola, and Alexei~A Efros.
\newblock {Unpaired Image-to-Image Translation using Cycle-Consistent Adversarial Networks}.
\newblock In \emph{Proceedings of the IEEE International Conference on Computer Vision (ICCV)}, 2017.

\bibitem[Zhu et~al.(2019)Zhu, Huang, Shi, Yu, Wang, and Bai]{zhu2019progressive}
Zhen Zhu, Tengteng Huang, Baoguang Shi, Miao Yu, Bofei Wang, and Xiang Bai.
\newblock {Progressive Pose Attention Transfer for Person Image Generation}.
\newblock In \emph{Proceedings of the IEEE Conference on Computer Vision and Pattern Recognition (CVPR)}, 2019.

\bibitem[Zwitserlood et~al.(2004)Zwitserlood, Verlinden, Ros, and Van Der~Schoot]{zwitserlood2004synthetic}
Inge Zwitserlood, Margriet Verlinden, Johan Ros, and Sanny Van Der~Schoot.
\newblock {Synthetic Signing for the Deaf: Esign}.
\newblock In \emph{Proceedings of the Conference and Workshop on Assistive Technologies for Vision and Hearing Impairment (CVHI)}, 2004.

\end{thebibliography}
}

\clearpage
\setcounter{page}{1}
\setcounter{section}{0}
\setcounter{footnote}{0}

\startcontents

\section{Overview of Supplementary Materials}

Below we provide more details, experimental results, and discussion. 
More details are in the \href{https://signllm.github.io}{https://signllm.github.io} project page.

\hypersetup{
     linkcolor=black
}

\printcontents{}{1}{}



\hypersetup{
     linkcolor=cvprblue
}
\begin{table*}[h]
    \centering
    \begin{tabularx}{0.95\textwidth}{|X|}
    \hline
        \textbf{Prompt Template \& Some Examples} \\ \hline
 \textbf{Part I}\\
I really want to learn how to say `\{\texttt{Text}\}' in sign language. Can you help me?\\
How would you express `\{\texttt{Text}\}' in sign language?\\
Can you show me how to say `\{\texttt{Text}\}' in sign language?\\
How do I say `\{\texttt{Text}\}' in sign language?\\ 
I am interested in mastering the sign language for `\{\texttt{Text}\}'.\\
What's the method to sign `\{\texttt{Text}\}'?\\
Can you show me how `\{\texttt{Text}\}' appears in sign language?\\
Could you tell me how `\{\texttt{Text}\}' is represented in sign language?\\
  \\ \hline
 \textbf{Part II}\\
How is ``So we're going to go up and down; let's switch hands, down and up; down and up." denoted in sign language?\\
Can you elucidate how And just let those fingers relax. looks in sign language?\\
Can you elucidate how `You do a full knot with both strands or a square knot with that.' materializes in sign language?\\
How do I say And I also use memory wire. with sign language?\\
I really want to learn how Now together you're going to go opposite. is said in sign language. Can you help?\\
How do I articulate ``It's real easy to actually get your fingers to lead, so try not to let them do that." using sign language?\\
I am intrigued to learn the sign language for `Let the wrist do all the leading.'.\\
I am wondering how ``Don't let the fingers take over, let the wrist do all the guiding." appears in sign language. \\
  \\ \hline
 \textbf{Part III}\\
Ich möchte wirklich lernen wie man `\{\texttt{Text}\}' in Gebärdensprache sagt. Können Sie mir helfen?\\
Wie würden Sie `\{\texttt{Text}\}' in Gebärdensprache ausdrücken?\\
Können Sie mir zeigen wie man `\{\texttt{Text}\}' mit Gebärdensprache sagt?\\
Wie sage ich `\{\texttt{Text}\}' in Gebärdensprache?\\
Könnten Sie mir sagen wie `\{\texttt{Text}\}' in Gebärdensprache dargestellt wird?\\
Mich interessiert wie man `\{\texttt{Text}\}' in Gebärdensprache sagt.\\
Können Sie die Gebärdensprache für `\{\texttt{Text}\}' demonstrieren?\\
Ich möchte erfahren wie `\{\texttt{Text}\}' in Gebärdensprache übersetzt wird.\\
  \\ \hline
 \textbf{Part IV}\\
`regen und schnee lassen an den alpen in der nacht nach im norden und nordosten fallen hier und da schauer sonst ist das klar' Wie stellt man das in Gebärdensprache dar?\\
Können Sie die Gebärdensprache für 'am donnerstag regen in der nordhälfte in der südhälfte mal sonne mal wolken ähnliches wetter dann auch am freitag' demonstrieren?\\
Mich interessiert, wie vom nordmeer zieht ein kräftiges tief heran und bringt uns ab den morgenstunden heftige schneefälle zum teil auch gefrierenden regen in Gebärdensprache aussieht.\\
Wie wird sonnig geht es auch ins wochenende samstag ein herrlicher tag mit temperaturen bis siebzehn grad hier im westen in Gebärdensprache dargestellt?\\
Wie würden Sie deutschland liegt morgen unter hochdruckeinfluss der die wolken weitgehend vertreibt gebärden?\\
Können Sie mir zeigen, wie am sonntag im nordwesten eine mischung aus sonne und wolken mit einigen zum teil gewittrigen schauern in Gebärdensprache aussieht?\\
Wie sieht die Gebärdensprache für örtlich schauer oder gewitter die heftig sein können aus?\\
Was ist die Gebärdensprache für und zum wochenende wird es dann sogar wieder ein bisschen kälter?\\
Was ist die Gebärdensprache für in der südhälfte weht der wind schwach sonst schwach bis mäßig richtung küsten frisch und stark böig?\\
  \\ \hline
    \end{tabularx}
    \caption{We provide two templates for sign language as a reference, and \texttt{\{Text\}} is where the video oral dialogue is inserted.}
    \label{tab:temp_supp}
\end{table*}
\begin{table*}[t]
\centering
\renewcommand{\arraystretch}{1.2}
\resizebox{\textwidth}{!}{
\begin{tabular}{lccccccccccccc} 
    \hline
    \midrule
        \textbf{Name}                            & \textbf{Language} & \textbf{Vocab.} & \textbf{Duration (h)} & \textbf{Signers} & \textbf{Multiview} & \textbf{Transcription} & \textbf{Gloss} & \textbf{Pose} & \textbf{Depth} & \textbf{Speech} & \textbf{Prompt} & \textbf{Compress}\\
    \hline                          
        Video-Based CSL~\cite{video-based}       & CSL               & 178       & 100           & 50 &\xmark & \greencheck & \xmark     & \greencheck     & \greencheck   & \xmark  & \xmark & \xmark\\
        SIGNUM~\cite{SIGNUM}                     & GSL               & 450       & 55            & 25 &\xmark     & \greencheck & \greencheck & \xmark     & \xmark   & \xmark & \xmark & \xmark \\
        RWTH-Phoenix-2014T~\cite{SLTranslation}  & GSL               & 3k        & 11            & 9 &\xmark     & \greencheck & \greencheck & \xmark     & \xmark   & \xmark & \xmark & \xmark \\
        Public DGS Corpus~\cite{DGS-Korpus}      & GSL               & --        & 50            & 327 &\greencheck & \greencheck & \greencheck & \greencheck & \xmark   & \xmark & \xmark & \xmark \\
        BSL Corpus~\cite{bsl-corpus}             & BSL               & 5k        & --            & 249 &\xmark     & \greencheck & \greencheck & \xmark     & \xmark   & \xmark & \xmark & \xmark \\ 
    \hline
        NCSLGR~\cite{NCSLGR}                     & ASL               & 1.8k     & 5.3            & 4 &\greencheck & \greencheck & \greencheck & \xmark     & \xmark     & \xmark & \xmark & \xmark \\
        How2Sign \cite{duarte2021how2sign}                 & ASL               & 16k       & 79            & 11 &\greencheck & \greencheck & \greencheck & \greencheck & \greencheck & \greencheck & \xmark & \xmark \\
    \hline
       \rowcolor{backgroundblue} \textbf{Prompt2Sign (ours)}                 & Multilingual               & 40k       & 200            & 40 &\greencheck & \greencheck & \greencheck & \greencheck & \greencheck & \greencheck & \greencheck & \greencheck \\
    \hline
    \midrule
\end{tabular}
}
\vspace{-4pt}
\caption{\textbf{Dataset Details:}
\ourDataName{} uses tools to automate the acquisition and processing of sign language videos on the web, is a better dataset that is efficient (a higher level of preprocessing, standardized and more models available), and lightweight (average reduction of 80\% in space usage).
Languages included: American Sign Language (ASL), German Sign Language (GSL, Alias DGS), Swiss German Sign Language (DSGS), French Sign Language of Switzerland (LSF-CH), Italian Sign Language of Switzerland (LIS-CH), Argentine Sign Language (Lengua de Señas Argentina, LSA), Korean Sign Language (KSL), and Turkish Sign Language (TSL).
}\label{tab:related_datasets_all}
\vspace{-5pt}
\end{table*}

\section{Background Information}

Here we expand on some of the nouns mentioned briefly:

\vspace{-12pt}
\paragraph{Gloss:} In the context of sign language, gloss refers to the process of providing a word-for-word translation of sign language into written or spoken language. It involves assigning a specific written or spoken word to each sign in order to facilitate communication and understanding between sign-language users and non-sign-language users. It generally represents a specific gesture or posture.


\vspace{-12pt}
\paragraph{OpenPose:} 
OpenPose\footnote{\href{https://github.com/CMU-Perceptual-Computing-Lab/openpose}{https://github.com/CMU-Perceptual-Computing-Lab/openpose}} is a real-time multi-person keypoint detection library that uses computer vision techniques to identify and track human body movements. The output result is a video of the key point visualization and key point data stored in json format for 24 frames a second.

\vspace{-12pt}
\paragraph{DensePose:} DensePose\footnote{\href{https://github.com/facebookresearch/detectron2/tree/main/projects/DensePose}{https://github.com/facebookresearch/detectron2/tree/DensePose}} is a method that estimates dense correspondences between a 2D image and a 3D human model. It can be used to extract detailed information about the body posture, position, and movements of sign language users from 2D images or videos, stored or displayed as a dense map covering the entire body of a human being. Details can be found in the footnote links.





\subsection{More Related Work}

Here, we introduce the third step of sign language production: Pose2Video, which involves visualizing key points in a video rendering or converting it into a live person/model demonstration of sign language. We also give some basic concepts of RL for a better understanding.

\vspace{-12pt}
\paragraph{Rendering of Conditional Input.}

Conditioning refers to the capacity of a generative model to manipulate its output based on our intentions. Previous instances of conditional input \acfp{gan} \cite{goodfellow2014generative} have exhibited favorable performance in generating images \cite{isola2017image,radford2015unsupervised,wang2018high,zhu2017unpaired} and videos \cite{mallya2020world,tulyakov2018mocogan,vondrick2016generating,wang2019few,wang2018video}. Numerous studies have also focused on generating human poses while considering various factors, including entire body \cite{balakrishnan2018synthesizing,ma2017pose,men2020controllable,siarohin2018deformable,tang2020xinggan,zhu2019progressive,chan2019everybody}, face \cite{deng2020disentangled,kowalski2020config,zakharov2019few,10.1007/978-3-030-58517-4_42,8995571,9747380}, and hand \cite{liu2019gesture,tang2018gesturegan,wu2020mm,fang2025signxfoundationmodelsign}.
One particular application is human-style transfer \cite{NEURIPS2022_ec795aea}, which involves replacing a person in a video with another individual while preserving their actions. This technique has also found extensive use in sign language production \cite{chan2019everybody,wei2020gac,zhou2019dance}. The key aspect lies in extracting keypoints to replicate movements \cite{chan2019everybody,ventura2020can}, utilizing tools such as OpenPose, i3D, and DensePose for common keypoint extraction \cite{chan2019everybody,wei2020gac,zhou2019dance,10.1145/3394171.3413532}.
In our work, we do not care about Pose2video, we only present some qualitative results at the end of the paper and in the supplementary materials.

\vspace{-12pt}
\paragraph{Reinforcement Learning.}

in the training or fine-tuning of large models is a common strategy. At the heart of reinforcement learning is the concept of a Markov Decision Process (MDP), an extension of Markov chains, which involves a finite set of states, a finite set of actions, state transition probabilities, and a reward function. The MDP delineates the interaction between an intelligent agent and the environment, wherein the agent chooses actions based on various states, and the environment imposes rewards or penalties on the agent based on the action and the current state, leading to a transition to the next state.
An optimal policy is the mapping from state $s$ to action $a$ that maximizes the total expected return:
\begin{equation}
\pi^* = \arg\max_{\pi} \mathbb{E}[G_t | s_t =s, \pi]
\end{equation}
where $G_t = \sum_{k=0}^{\infty}\gamma^k R_{t+k+1}$, $0 \leq \gamma < 1$ is the discount factor, and $\mathbb{E}[]$ is the expectation operator.
In LLMs, researchers often fine-tune models with reinforcement learning based on human feedback. Given that the \ac{slp} process aligns with the definition and can be reformed by the MDP, we simply simulate this concept to fine-tune our generation model. However, since the training scenario of sign language does not involve interaction with the environment, our reinforcement learning strategy is not a typical one, but rather only partially applied to component modules.

\section{More Details of Prompt2Sign}\label{sec:more_prompt2sign}

\subsection{Dataset Modalities}
\label{subsec:content}

In comparison to the previous datasets, we possess numerous additional advantageous attributes and a larger scale. As with the previous dataset work, we extracted everything automatically except speech/text. But we've added some automated channel tools that go deeper than that.

\vspace{-12pt}
\paragraph{Prompt Word Templates.}
\label{subsec:templates}
We constructed 120 English templates and 210 prompt word templates generated by GPT4 for other languages (with 30 templates for each language), which were randomly associated with the script data to form a part of our dataset. Some examples are in Table \ref{tab:temp_supp} above, our prompt templates are designed to ask how to demonstrate sign language, making it easier for LLMs to recognize users' sign language queries. For the simple question ``how to demonstrate `xx'", 30 templates have covered 99\% of use cases, making it unlikely to obtain more without overlap. This prompt word data is needed for the future development of large language models of sign language because we were able to develop a model with understands more complex, natural human conversational inputs by using prompt word data.

\vspace{-12pt}
\paragraph{Data Enhancement.} \label{paragraph:data_enhancement} 
In our multilingual sign language production tasks, we found that the model underperformed in low-resource sign languages. For these sign languages, where it's inherently difficult to increase the data volume, we use tools to rewrite lines or prompt words and can obtain several times more data to enhance the robustness of the trained model in low-resource data.

\vspace{-12pt}
\paragraph{Multiview.} Our multiple perspectives depend on the original video, and it is worth noting that if the researchers cannot guarantee that the newly acquired perspectives are all positive, then the model will generally be contaminated.

\vspace{-12pt}
\paragraph{Depth Data.}
Our depth depends on whether the raw data video has relevant support, we believe that this is generally not needed, as most work uses lifting work to obtain 3D key points, rather than high-cost professional equipment.




\subsection{Pose Information}

\paragraph{Necessity of Uniform Standards.}
If there is a mismatch between any of these components in \ac{slp} \cite{wang2018video,wang2018high,chan2019everybody,saunders2020progressive,stoll2020text2sign,saunders2021mixed} or SLT \cite{camgoz2018neural,camgoz2020sign,ko2019neural,Bohacek_2022_WACV}, it can lead to complex challenges. For instance, if the results of pose recognition cannot be used as training data, the results of SLR cannot be used for model testing, or if the results of sign language generation cannot be used as conditional input \cite{chan2019everybody,wei2020gac,zhou2019dance}, which have troubled many novice researchers in the field.

\vspace{-12pt}
\paragraph{Data Format Conversion.}

\begin{itemize}
\item \textbf{How to extract key points?} We extracted two-dimensional (2D) frontal human pose information from videos of different resolutions, including upper body pose information of the body and hands, through OpenPose~\cite{openpose}. Includes 8 upper body key points. 21 keypoints in each hand, which is a total of 42 hand keypoints. These two parts add up to fifty keypoints, each of which has three XYZ messages, or 150 numbers.
\end{itemize}
Then in steps ``json (2D keypoints) to h5'', ``h5 to txt (3D keypoints)'', and ``txt to compressed data'':

\begin{itemize}
\item \textbf{How to complete ``json to h5''?} We successively obtain a json number in a folder (a frame of pose information, 50 key points, 150 numbers), and then read all the json numbers in a folder into the key name of an h5 (h5 is a format of numpy) file, multiple folders form multiple build names, and finally form an h5 file.
\end{itemize}

\begin{itemize}
\item \textbf{How to complete ``h5 to txt''?} We read each key name of h5 in turn (the original folder name), create the corresponding folder, each folder generates 5 txt files, the last one is the result, the first 4 txt stores the intermediate variable. This is the part of 2D to 3D, and the key formula 3 in the text is the formula of this part. Additionally, we read the relevant data and delete the unqualified data such as NaN, 0, or replace it with the average median. Finally, we condensed the data to about 1/5 of the original, which comes from the processing of the ASL part.
\end{itemize}

\begin{itemize}
\item \textbf{How to complete ``txt to compressed data''?} We read the fifth txt file of each folder in turn, the number of lines in the txt file represents the number of frames of the folder corresponding to the video. We read a line of txt (150 numbers, separated by Spaces, a frame of information), plus a space, and then add a count value (the current line divided by the total number of lines, representing the progress bar), add a space after the count value, Then add the second line txt and continue to repeat the above. Then we put a txt (video information, the total number of numbers in it = 151 $\times$ video frames) into a line of content, in turn, tens of thousands of videos are all stored in our standard format.
\end{itemize}


\subsection{More Details of the Data} \label{subsec:3-Steps_Tools}

\paragraph{Details of Processing.}
Firstly, we obtain the original video from the internet. As mentioned in the main text, this part still needs to be done manually, but a script can be written to speed up the process. 
Firstly, preliminary preprocessing can be done through scripts written by oneself or OpenASL \cite{shi-etal-2022-openASL} scripts
. Secondly, the dialogue of the video is transcribed into text, videos are processed using OpenPose, and then used as input for our tool. Finally, enters the language mode corresponding to the data by setting the model to start training.

\vspace{-12pt}
\paragraph{Time and Cost of Dataset Processing}
Among all the data processing steps, the most time-consuming step is 2Dto3D, 
a GPU can process 1000 clips after 10 hours, and can process 50-80 hours of How2Sign data in about half a month (there is no 80 after editing). Improving the performance of a single card does not make it much faster, which may be caused by multithreading concurrency restrictions.







\begin{figure}[t]

\DeclareRobustCommand*{\RaiseBoxByDepth}{%
    \raisebox{-0.2\height}%
}

\vspace{-0.05cm}
    \centering
    \includegraphics[width=\linewidth]{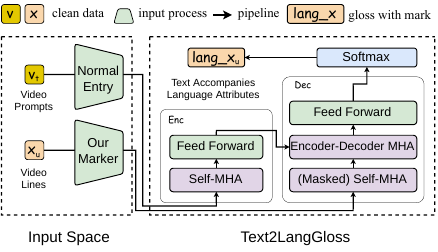}
    \vspace{-18pt}
    \caption{We enhance Text2Gloss \cite{saunders2020progressive} with a marker to generate the Gloss with linguistic properties. The $\bm{v}_{t}$ (\RaiseBoxByDepth{\includegraphics[height=8pt]{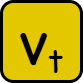}}) and $\bm{x}_{u}$ (\RaiseBoxByDepth{\includegraphics[height=8pt]{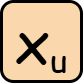}}) represent data types and abstract representations.}
    \label{fig:prompt2langloss}
    \vspace{-12pt}
\end{figure}%

\subsection{Prompt2LangGloss.}\label{sec:text_to_langgloss}

As shown in Figure~\ref{fig:Model_Overview} (Left), our proposed enhancement of this model involves appending an additional language attribute to each Text word during the reading and tokenizing stages. For instance, a traditional gloss token ``\texttt{<xxx>}" can be transformed into ``\texttt{<ASL\_xxx>}", thus introducing a layer of conditional input $f_{u} = E_{T2LG}(x_{u} | x_{1:U})$ into \ac{slp} based on Eq. \ref{eq:T2P_encoder}:
$lg_{w+1} = D_{T2LG}(lg_{w}  | lg_{1:w-1} , f_{1:\mathcal{U}})$.
This figure is a supplement to the main paper text.

During this process, we can see in detail how our components operate and what the encoder-decoder structure is. In the main text, we noticed that more complex prompt words act as noise relative to the text we actually want to translate. Therefore, we conducted some experiments in Table \ref{tab:prompt} to verify whether this impact could be reduced or eliminated.




\section{More Experiments}\label{sec:more_experiments}



\subsection{Extensibility \& Visual Study} \label{paragraph:more_Visual_Study}

Subsequently, we provide an overview and comparison of motion capture techniques and novel visual models. Our objective is to advocate for adopting motion capture technology as a replacement for traditional visual methods in sign language rendering; they can reduce the finger-missing problems mentioned in the main text of the paper. Before that, we need to introduce some background:

\vspace{-8pt}
\subsubsection{Motion \& Visual Method Introduction} 


\paragraph{SMPL skeleton system:} The SMPL \cite{smpl} (Skinned Multi-Person Linear) skeleton system is a parametric model that represents human body shape and pose. It is commonly used in computer graphics and animation. In the context of sign language, the SMPL skeleton system can be utilized to model and animate sign language movements and gestures. 

\vspace{-8pt}
\paragraph{VMD files and OpenMMD:} VMD (Vocaloid Motion Data) files and OpenMMD (Open-source MikuMikuDance) refer to specific file formats and software tools used in character animation. VMD files contain motion data and are commonly used in the MikuMikuDance software for animating virtual characters.
OpenMMD is an open-source implementation that allows users to create and modify character animations. 
In the context of sign language, VMD files and OpenMMD can be utilized to animate virtual characters performing sign language gestures or movements.

\vspace{-8pt}
\paragraph{Keypoint driven model:} A key point driven model is a computational model or algorithm that relies on the detection and tracking of specific key points, landmarks, or features in order to analyze and interpret data or generate desired outputs. 
In the final pose-to-video stage of sign language rendering, the generation of realistic human videos from keypoints is essential. This can be accomplished through either motion capture or purely visual methods. In the following sections, we will evaluate the strengths and limitations of each approach. In the context of sign language, a keypoints-driven model can be used to analyze and interpret sign language movements based on the detection and tracking of key points on the signer's body, such as hand positions, facial expressions, and body postures. It is our tentative exploration in this work.

\vfill\break

\begin{figure}[ht]
    \centering
    \footnotesize
    \includegraphics[width=\linewidth]{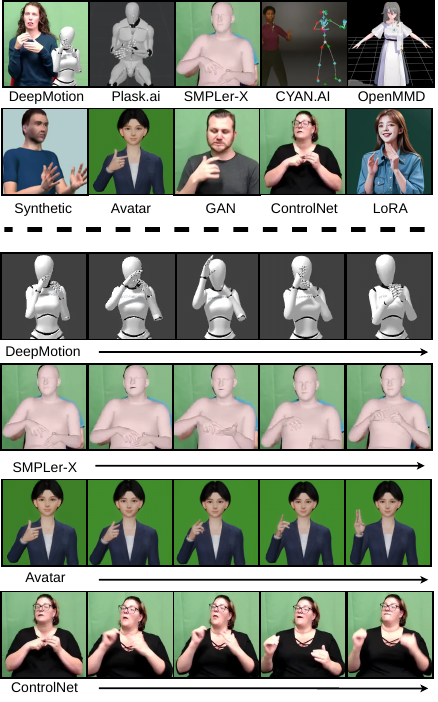}
    \vspace{-18pt}
    \caption{\textbf{Extensibility Presentation:} We used five motion capture models and five sign language rendering models to show the final production effect.}
    \vspace{-14pt}
    \label{fig:expand_supp}
\end{figure}

\subsubsection{Comparison of Motion and Visual} 

\paragraph{Extensibility Study.}

In Figure \ref{fig:expand_supp}, the first line of it is obtained either directly or indirectly by reading our \ourLLMmethodName{} output sequence through motion capture\footnote{\href{https://portal.deepmotion.com/}{DeepMotion}; \href{https://plask.ai/}{Plask.ai}; \href{https://avatar.aliyun.com/}{Avatar}; \href{https://github.com/peterljq/OpenMMD}{OpenMMD}} \cite{chen2023executing,cai2023smplerx} software or models, while the second line of the image comes from the commonly used Pose2Vid \cite{mallya2020world,tulyakov2018mocogan,vondrick2016generating,Rombach_2022_CVPR,zhang2023adding,hu2021lora} or Pose2Img \cite{isola2017image,radford2015unsupervised,wang2018high,zhu2017unpaired} models. The broad scope of our model becomes apparent from the initial two statements. Subsequently, the next four lines present sign language demonstration videos created using either direct or indirect input of keypoints (some videos sourced from the project website). It is important to note that SMPLer-X and Avatar are utilized solely for demonstrative purposes in this context. Taking DeepMotion and VMD as instances, our model exhibits the capability to operate within a broader scope by utilizing keypoints as input, rather than relying solely on visual methods. This advancement provides the potential for more precise sign-language demonstrations. Details can be found in the footnote links.


\begin{table}[t]
\centering
\resizebox{0.99\linewidth}{!}{%
\begin{tabular}{@{}p{3.0cm}ccccc@{}}
\toprule
  & \multicolumn{1}{c}{SSIM $\uparrow$} & \multicolumn{1}{c}{Hand SSIM $\uparrow$} & \multicolumn{1}{c}{Similarity $\uparrow$} & \multicolumn{1}{c}{F2FD $\downarrow$} \\ \midrule
\multicolumn{1}{r|}{Vid2Vid \cite{wang2018video}} & 0.743 & 0.582 & 78.42 & 27.86 \\
\multicolumn{1}{r|}{ControlNet \cite{zhang2023adding}} & 0.817 & 0.646 & 82.11 & 25.47 \\
\multicolumn{1}{r|}{Motion Capture} & \textbf{0.826} & \textbf{0.687} & \textbf{81.29} & \textbf{22.71} \\
\bottomrule
\end{tabular}%
}
\vspace{-1mm}
\caption{\textbf{Visual Study:} SSIM: Comparison of image structure similarity between the generated image and the condition graph extracted from the Ground Truth. Similarity: Extract the similarity percentage of keypoints between the generated video and the input action. F2FD: The degree of difference between frames.}
\label{tab:visual_supp}
\vspace{-12pt}
\end{table}

\vspace{-8pt}
\paragraph{Visual Study.}
We explored the influence of different forms on performance as shown in Table \ref{tab:visual_supp}, current existing motion capture models do not fully support our keypoints format, and there may be some loss in certain transmission processes. Therefore, our primary focus is evaluating the presentation effect of motion capture models in sign language. Taking DeepMotion as an example, it is a deep learning-based method that drives models in a software environment using keypoints. In previous work, the comparison between rendered results and GroundTruth was measured using the structural similarity index (SSIM). However, since driving models do not have a specific GroundTruth, our comparison is based on the visualized keypoints extracted, which may introduce some errors but generally remain below 1\%, providing a sufficient basis for simple comparisons. The percentage similarity refers to the comparison of extracted sequence numbers. Additionally, the difference between frames focuses on the smoothness of the video, as motion capture models do not exhibit the flickering issue common in generative models, resulting in smaller differences between consecutive frames. While the software can output a higher number of frames for enhanced results, we set the frame rate to 24 frames per second for fair comparisons. In conclusion, we believe that introducing motion capture-related techniques, models, or methods holds great promise in the final rendering stage of sign language. 

\subsection{Model Parameter Study}
%
As shown in Table \ref{tab:ablation_parameters}, we have investigated the optimal parameter settings under different circumstances to provide further discussion and guide future researchers in their training. This includes the optimal results of our primary model parameters, architecture, and various learning rates or other parameters. The experimental results were derived by evaluating the performance of the Text to Pose function from the \ourLLMmethodName{}-40M-Base model to the \ourLLMmethodName{}-1B-Large model on the ASL part of \ourDataName{} dataset.
In general, we find that (1) The optimal values of the parameters conform to the scaling law \cite{Hoffmann-arxiv-2022-Training, Kaplan-arxiv-2020-Scaling}, according to which we should increase the number of parameters by four times when the data is increased by four times. 
There is no significant difference between the 120M model and the 40M model prediction without too much increase in data volume, and there is also a larger magnitude. (2) When we use Prompt2LangGloss, our data equals the sum of two single language versions. But at this time, their performance mainly depends on the data of a single language, which is a special case: Although they share parameters, the LangGloss has distinguished enough of the sign language pose corresponding to the input text, they do not enjoy the bonus of shared parameters.
\begin{table}[t]
\centering
\resizebox{1\linewidth}{!}{
    \begin{tabular}{lll}
    \toprule
    Key name              & Values    & Note            \\ \midrule
    BSLP method        & \{\underline{\textbf{ASL}}, GSL\}    & Choose before training         \\
    Vocabulary size            & \{1k, 4k, 7k, \underline{\textbf{16k}}\}  & Case-sensitive         \\ \midrule
    Batch size                 & \{\textbf{8}, \underline{16}, 32\}       & Adjust according to configuration      \\
    Learning Rate (LR)\tablefootnote{LR is getting smaller and smaller over time, approaching a set value.}
    & \{5e-2, \underline{\textbf{1e-3}}, 5e-3\} & Training initial value  \\
    Loss mode               & \{\underline{\textbf{MSE}}, RL, L1, L2, LV\}  & Adjust according to situation \\
    Max\_sent\_length             & \{\underline{300}, \textbf{400}\}        & Input is usually less than maximum     \\
    Priority Learning Channel           & \{\underline{False}, \textbf{True}\}     & Use with RL Loss        \\
    Dropout                    & \{\underline{\textbf{0}}, 0.1, 0.2, 0.3\}  & Adjust according to situation \\
    \# Layers (encoder-decoder) & \{2-2, \underline{4-4}, \textbf{8-8}\} & Not necessarily correspond
 \\
    Embed dim                  & \{512, \underline{1024}, \textbf{2048}\}   & Adjust for the amount of data         \\
    FFN dim                    & \{2048, \underline{4096}, \textbf{8192}\}  & Must equal to 4*hidden size   \\
    \# Attention heads        & \{4, \underline{8}, \textbf{16}\}          & Adjust for the amount of data     \\ \bottomrule
    \end{tabular}
}
\vspace{-4pt}
\caption{\textbf{Hyperparameters Space:} Optimal choices are marked in \textbf{bold}, while defaults are \underline{underlined}. The default values come from the \ourLLMmethodName{}-120M-Base-M (ASL).}
\label{tab:ablation_parameters}
\vspace{-5pt}
\end{table}

\begin{table}[t!]
\centering
\resizebox{0.99\linewidth}{!}{%
\begin{tabular}{@{}p{3.0cm}ccccc@{}}
\toprule
     & \multicolumn{2}{c}{DEV SET}  & \multicolumn{2}{c}{TEST SET} \\
\multicolumn{1}{c}{Approach:}  & \multicolumn{1}{c}{BLEU-4 $\uparrow$} & \multicolumn{1}{c}{ROUGE $\uparrow$} & \multicolumn{1}{c}{BLEU-4 $\uparrow$} & \multicolumn{1}{c}{ROUGE $\uparrow$} \\ \midrule
\multicolumn{1}{r|}{Stoll \etal \cite{stoll2018sign}} & 16.34 & 48.42 & 15.26 & 48.10 \\
\multicolumn{1}{r|}{Baseline \cite{saunders2020progressive}} & 20.23 & 55.41 & 19.10 & 54.55 \\
\multicolumn{1}{r|}{\textbf{Ours}} & \textbf{23.10} & \textbf{58.76} & \textbf{22.05} & \textbf{56.46} \\
\multicolumn{1}{r|}{\textbf{$\Delta$ \textcolor{sgreen}{$Acc.$}}} & \textcolor{sgreen}{\textit{\textbf{+ 14.2\%}}} & \textcolor{sgreen}{\textit{\textbf{+ 6.0\%}}} & \textcolor{sgreen}{\textit{\textbf{+ 
 15.4\%}}} & \textcolor{sgreen}{\textit{\textbf{+ 3.5\%}}} \\
\bottomrule
\end{tabular}%
}
\vspace{-4pt}
\caption{\textbf{Prompt Channel Accuracy:} We investigate the Prompt2LangGloss channel information loss by German SLP.}
\label{tab:prompt}
\vspace{-5pt}
\end{table}

\vspace{-8pt}
\paragraph{Prompt Fine-Tuning and User Study.} \label{paragraph:prompt_study}
By employing prompt words as input for the text channel and using the original text or the original gloss as input for the gloss channel (\ie, usage of Prompt2LangGloss), we can develop a model with understanding prompts competency. This approach aims to translate natural language into objective text/gloss before inputting it into the model. In reality, users might question, \textit{``How do you demonstrate `the sky is blue' in sign language?''}, rather than directly inputting \textit{``the sky is blue''}. This training strategy gives the model a degree of Understanding more complex input text ability. We compared the Prompt2LangGloss channel with a tokenizer to the previous Text2Gloss approach in terms of prompt usage. The experiments indicated that the impact of the tokenizer in the Prompt2LangGloss channel is relatively small and can be overcome through better training. As presented in Table \ref{tab:prompt}, it underscores the effectiveness of our mode in reducing semantic information loss in the channel.

\section{More Discussion} \label{sec:more_discussion}

\paragraph{Discussion on The Dataset Range.} 
We have cited the sources of our publicly available data, and some of the more popular works were not considered due to their limited accessibility and potential usage restrictions. Additionally, while there are other multilingual datasets available \cite{gueuwou2023jwsign,yin2022mlslt,matthes2012dicta,hilzensauer2015multilingual}, they may not possess the same level of comprehensiveness as ours. Like \cite{gueuwou2023jwsign} and \cite{yin2022mlslt}, they are papers that translate two types of sign language videos into spoken language (SLT), while our work is from spoken language to videos (SLP). We aim to develop the multilingual SLP method, and our dataset has more diverse application scenarios than them (\eg, one is Bible translation\footnote{\href{https://aclanthology.org/2023.findings-emnlp.664/}{https://aclanthology.org/2023.findings-emnlp.664/}} \cite{gueuwou2023jwsign}, the other\footnote{\href{https://ieeexplore.ieee.org/document/9878501}{https://ieeexplore.ieee.org/document/9878501}} \cite{yin2022mlslt} is cross SLT, but we are comprehensive scenarios and SLP. Some are multilingual dictionary datasets \cite{matthes2012dicta,hilzensauer2015multilingual}). We are a very beneficial supplement to previous work, which is quite different from previous work.


\vspace{-8pt}
\paragraph{Discussion on The Dataset Errors.} 

We handle issues related to NaN, zero, and missing data by applying deletion or replacement techniques and our tool simplifies certain calibration stages in comparison to previous 2D to 3D tools, which may introduce some errors. The substituted data is derived using median or mean values, resulting in minuscule errors. Within the vast parts of dataset, these errors typically fall within the range of 0.5\% to 0.7\% (We conducted a random sampling of results and obtained a ratio of 87 out of 17,549 to 47 out of 6,685). Moreover, our processing steps involving normalization greatly diminish such errors. Hence, we have reasonable grounds to assert that the data error is minimal enough to be bearable. In addition, we consider reducing these errors before release. And some of the data sets have some potential problems, and the number of our data sets is based on the final release.

\vspace{-8pt}
\paragraph{Discussion on The Prompt2Pose Task?} 
If existing methods want to use long text prompts as input, they must obtain more data to achieve better results. However, sign language data is very scarce. We need to propose a new Prompt2LangGloss method that can efficiently translate based on the scarcity of sign language. Therefore, it is necessary to propose Prompt2Pose as an independent task, as this module has many application scenarios even in a wider range of action synthesis (\eg, the robot can understand human commands more efficiently, so that the LLM to summarize the Prompt is not required). Compared to using LLM to simplify and summarize everyone's input, creating an efficient end-to-end sign language model is very valuable, similar to the popular Retrieval Augmented Generation (RAG), but for sign language.

\begin{table*}[ht]
\centering
\resizebox{1\textwidth}{!}{
\begin{tabular}{c|c|c|c|c|c|c}
  \hline\hline
  Mode & Function & Address & Enc-Dec & Prompt & Feature & Note \\
  \hline
  $M$ & multilingual SLP & text2pose & Multiple & No & More efficient/stable & Language is easy to add or subtract \\
  \hline
  $P$ & multilingual SLP & text2gloss & Single & Allow & Understand complex input & Greater potential for development\\
  \hline
\end{tabular}
}
\vspace{-4pt}
\caption{\textbf{The difference between the two modes:} $M$ and $P$ represent MLSF and Prompt2LangGloss, respectively. Adress represents which traditional step has been innovated, while Features represents the ability of the mode to focus more on.}
\label{tab:different}
\vspace{-5pt}
\end{table*}

\vspace{-8pt}
\paragraph{Why do we need to have two models?} 
\noindent The reasons for designing two modes are as follows:

\begin{enumerate}
   \item \textbf{Different Usage Scenarios:}
   \begin{itemize}
       \item MLSF targets direct text translation needs
       \item Prompt2LangGloss addresses sign language-related Q\&A and instruction needs
   \end{itemize}

   \item \textbf{Research Purpose Level:}
   \begin{itemize}
       \item Provides two different paradigms for researchers to reference; Demonstrates two different approaches for multilingual sign language generation
       \item Proves that existing/LLM models can be transformed into multilingual models through different approaches; These two directions will be the mainstream approaches in the future
   \end{itemize}


   \item \textbf{Practical Significance:}
   \begin{itemize}
       \item Provides solutions for different application scenarios
       \item Enables sign language models to serve different needs more flexibly
   \end{itemize}
\end{enumerate}

Therefore, the core reason for using two separate modes is: they address different needs in sign language generation tasks while providing diverse technical paradigm references for the research community.

\vspace{-8pt}
\paragraph{Why can't the two modes merge into one?} 
Because the two modes cannot coexist in different AB stages. Let's imagine a scenario where you're going to a distant place, with two stages, A and B, occurring in sequence. If you have three paths to choose from in Stage A and only one path in Stage B, you will ultimately have 3 $\times$ 1 choices. If you have only 1 path in stage A and 3 paths in stage A, overall, you still have 1 $\times$ 3 choices. 

Therefore, they cannot coexist because 3 $\times$ 3 is a meaningless and more complex choice: If you want to implement 9 languages (9 choices), you only need to modify either the first stage or the second stage into nine paths. There's no need to modify both stages separately, as this would make the model unnecessarily complex. To implement multilingual sign language production in a model, only one stage of AB needs to develop multiple paths (languages).

\vspace{-8pt}
\paragraph{The difference of Two Modes} 

MLSF dynamically adds encoders, which can avoid semantic confusion and maximize its convenience (\eg, a general model can execute multilingual SLP tasks that were impossible for researchers in the past. It saves significant development time, potentially twice the effort, ten times the return). Prompt2LangGloss focuses on improving the ability to understand complex inputs, which is complementary to the MLSF. It will have great prospects with the data volume increase (\eg, ChatGPT rarely mixes languages when speaking in a specific language). Moreover, LangGloss can be used without choosing a language as a valuable feature. Therefore, both approaches have their focus, input type, and user cases, summarized in Table \ref{tab:different}.


\vspace{-8pt}
\paragraph{Discussion on Multilingual SLP Task.}
As mentioned in Table \ref{tab:mslp} of the main text, most lesser-known sign languages basically have no baseline data for sign language production. This may be due to various reasons, and some datasets don't even have baseline data for sign language recognition. In this situation, it's difficult for us to replicate work from many years ago, and it's challenging to obtain the performance level of sign language production models for these languages. Therefore, for transformer-based or deep learning-based sign language production, we indeed proposed the first baseline for these languages without needing to compare with any previous models (of course, in most cases, these models don't exist).

\vspace{-8pt}
\paragraph{Discussion on Devset Performance.}

Scholars may wonder why our model demonstrates higher performance on the development set compared to the test set, and why previous studies typically focused on test set evaluations rather than development set assessments. Our model's backbone is \href{https://github.com/BenSaunders27/ProgressiveTransformersSLP}{PT}. The code uses performance on development set to select the best checkpoint
This means the model parameters are optimized based on devset performance rather than test set. Therefore, it is normal for test set performance to be lower than dev set performance, even though neither dev nor test sets are involved in training. Many previous works have followed this path (not trained in devset, but optimized for it), if some works not follow this, we don't need to compare with them on devset.

\vspace{-8pt}
\paragraph{Discussion on Similar Name.} We noticed that a work \cite{gong2024llms} has introduced a large language model to translate sign language videos into spoken text. Their proposed framework is also called SignLLM. Since our work focuses on converting spoken text to sign language videos, which is the opposite direction, there is no conflict between the two approaches despite sharing the same name.

\vspace{-8pt}
\paragraph{Discussion on The Significance of Our Dataset.} 

Sign language is different from most motion recognition fields, requiring many complex annotations, and video datasets generally have a low level of processing. This work is also very time-consuming and labor-intensive, even more challenging than creating an original video dataset with more than a dozen languages. What we need is high-quality data. If each researcher needs to process thousands of hours of data before starting experiments, many researchers will lose enthusiasm, which happens frequently in this field. Therefore, our new Prompt2Sign work is very necessary.

\vspace{-8pt}
\paragraph{Discussion on Dataset Licence}
The referenced papers are the sources of our datasets. For datasets that don't allow redistribution, we plan to provide tools for researchers to process the data themselves (a common practice among many well-known datasets, eg, \href{https://github.com/EricGuo5513/HumanML3D#how-to-obtain-the-data}{HumanML3D}, \href{https://github.com/IDEA-Research/Motion-X#3-mocap-subsets}{Motion-X}).


\stopcontents

\end{document}